\documentclass[letterpaper]{article}
\usepackage{aaai2026}
\usepackage{times}
\usepackage{helvet}
\usepackage{courier}
\usepackage[hyphens]{url}
\usepackage{graphicx}
\usepackage{natbib}
\usepackage{caption}
\usepackage{algorithm}
\usepackage{algorithmic}
\usepackage{newfloat}
\usepackage{listings}
\DeclareCaptionStyle{ruled}{labelfont=normalfont,labelsep=colon,strut=off}
\floatstyle{ruled}
\newfloat{listing}{tb}{lst}{}
\floatname{listing}{Listing}

\title{Human Motion Unlearning}
\author {
    Edoardo De Matteis\equalcontrib\textsuperscript{\rm 1},
    Matteo Migliarini\equalcontrib\textsuperscript{\rm 1},
    Alessio Sampieri\textsuperscript{\rm 2},
    Indro Spinelli\textsuperscript{\rm 1},
    Fabio Galasso\textsuperscript{\rm 1}
}
\affiliations {
    \textsuperscript{\rm 1}Sapienza University of Rome\\
    \textsuperscript{\rm 2}ItalAI
}
\usepackage{bibentry}

\usepackage{subcaption}
\usepackage{amssymb}
\usepackage{amsmath}
\usepackage{booktabs}
\usepackage{xcolor}
\usepackage{dsfont}
\usepackage{multirow}
\usepackage{tabularx}
\usepackage{pifont}
\newcommand{\ColorTable}[0]{666666}
\newcommand{\MMnotox}[0]{MM-Safe}
\newcommand{\cmark}{\ding{51}}
\newcommand{\xmark}{\ding{55}}
\newcolumntype{Y}{>{\centering\arraybackslash}X}
\begin{document}
\maketitle

% Abstract
\begin{abstract}
We introduce Human Motion Unlearning and motivate it through the concrete task of preventing violent 3D motion synthesis, an important safety requirement given that popular text-to-motion datasets (HumanML3D and Motion-X) contain from 7\% to 15\% violent sequences spanning both atomic gestures (e.g., a single punch) and highly compositional actions (e.g., loading and swinging a leg to kick). By focusing on violence unlearning, we demonstrate how removing a challenging, multifaceted concept can serve as a proxy for the broader capability of motion ``forgetting.''
To enable systematic evaluation of Human Motion Unlearning, we establish the first motion unlearning benchmark by automatically filtering HumanML3D and Motion-X datasets to create distinct forget sets (violent motions) and retain sets (safe motions). We introduce evaluation metrics tailored to sequential unlearning, measuring both suppression efficacy and the preservation of realism and smooth transitions.
We adapt two state-of-the-art, training-free image unlearning methods (UCE and RECE) to leading text-to-motion architectures (MoMask and BAMM), and propose Latent Code Replacement (LCR), a novel, training-free approach that identifies violent codes in a discrete codebook representation and substitutes them with safe alternatives.
Our experiments show that unlearning violent motions is indeed feasible and that acting on latent codes strikes the best trade-off between violence suppression and preserving overall motion quality. This work establishes a foundation for advancing safe motion synthesis across diverse applications. 
% Project page: \url{https://www.pinlab.org/hmu}.
\end{abstract}

\begin{links}
\link{Website}{https://www.pinlab.org/hmu}
\end{links}

% Introduction
\section{Introduction}

\begin{figure}[!t]
    \includegraphics[width=0.4\textwidth]{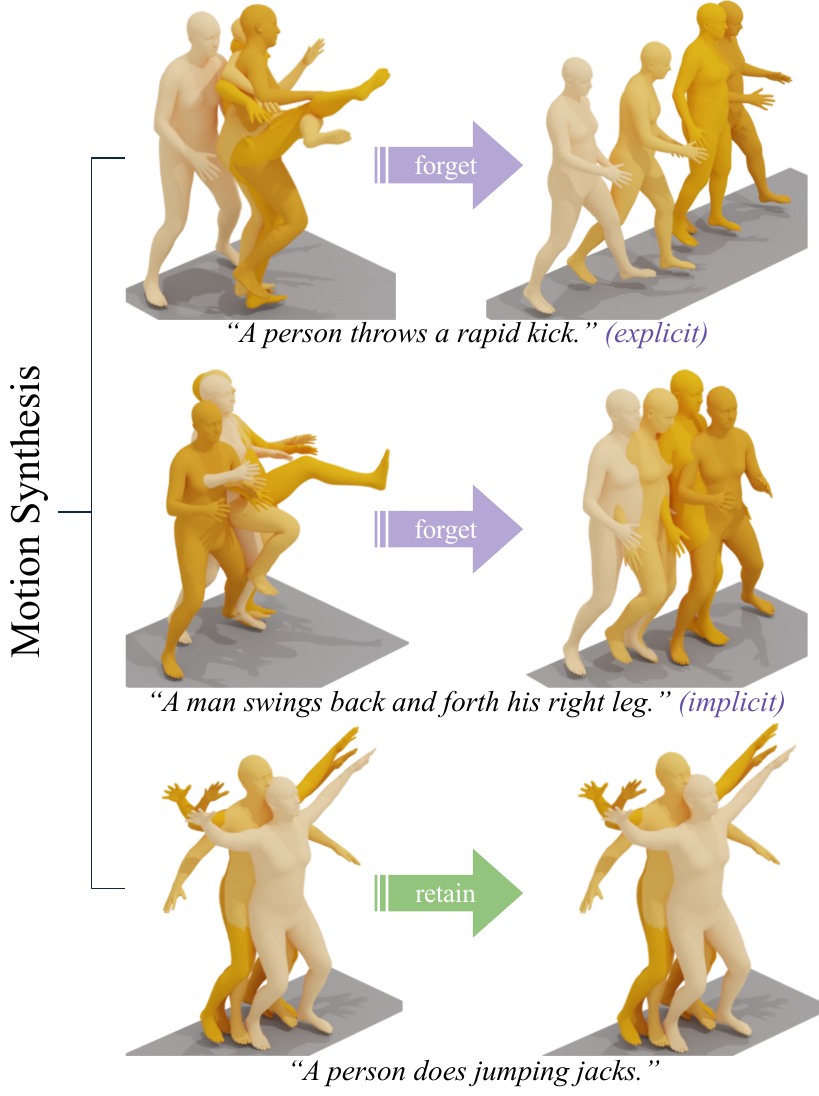}
    \caption{ The text-to-motion model takes an input prompt and generates the corresponding motion. With unlearning, when violent content is prompted, the model avoids generating harmful actions, producing a safe and appropriate outcome.}
    \label{fig:teaser}
\end{figure}

Generative models have seen dramatic progress across multiple modalities: images~\cite{Rombach2021HighResolutionIS, Ruiz2022DreamBoothFT}, videos~\cite{Fei2023DysenVDMED, blattmann2023videoldm}, music~\cite{copet2023musicgen}, and, more recently, 3D human motions~\cite{zhang2023t2mgpt, chen2023mld, sampieri24ladiff, jiang2023motiongpt}. Text-to-motion (T2M) systems now offer unprecedented realism and control, enabling applications in virtual reality~\cite{tirinzoni2024zero}, animation, and the development of embodied agents and robots trained on human motion data~\cite{pyroki2025, shao2025langwbclanguagedirectedhumanoidwholebody, serifi2024robot}. However, since popular training corpora such as HumanML3D~\cite{humanml3d} and Motion-X~\cite{motionx} encode examples of violent actions (e.g., punching, kicking, stomping), current T2M models can reproduce aggressive behaviors on demand, raising serious safety and ethical concerns. Uncontrolled generation of violent motions risks misuse in simulated environments, flawed human–robot interactions, and unintended bias in downstream controllers that inherit such behaviors~\cite{xu23survey1, merel2017learning, wang2025expertsgeneralistgeneralwholebody}. 

To address this gap, we introduce the task of Human Motion Unlearning (HMU), with a focus on violence unlearning. 
Violence serves as an ideal case study as it spans atomic gestures (e.g., a single punch) to highly compositional sequences (e.g., a kick followed by a grapple), requiring fine-grained suppression without degrading non-violent sub-motions. 
Beyond its role in mitigating real-world risks, such as preventing the synthesis of harmful behaviors in animation and robotics~\cite{merel2017learning, Yan23imitationnet, wang2025expertsgeneralistgeneralwholebody}, violence unlearning also provides a stringent benchmark for motion unlearning more broadly, demonstrating the feasibility of removing specific concepts from a trained generative model.

While we recognize that violent motions are legitimate for specific professionals (e.g., animators), our work focuses specifically on the safety of open models. We address the risks of generating harmful content when these models are distributed to the public, particularly in sensitive fields like robotics. We select violence as our case study because it is both prevalent and conceptually complex in existing datasets, making it an effective benchmark for a safety-oriented unlearning task. Other sensitive motions are too rare for systematic evaluation. Crucially, our approach is not a blunt filtering mechanism: since unlearning operates at the parameter level, it enables developers to release a “safe” public version of a model while retaining a full, unrestricted version for professional use.

We propose a dedicated benchmark for violence-free motion generation based on two recent, large-scale datasets: HumanML3D and Motion-X. From each corpus, we filter out sequences annotated with aggressive movements (punching, kicking, beating, stabbing, etc.) to produce a violence-free subset for evaluation. To capture the sequential nature of motion, our benchmark also includes a violent-only subset and a suite of motion-unlearning metrics that assess both the degree of violence suppression and the preservation of realism and smoothness in transitions between censored and uncensored segments.

Building on state-of-the-art (SotA) image unlearning techniques, we adapt two training-free methods, UCE~\cite{gandikota2024uce} and RECE~\cite{gong2024rece}, to leading T2M architectures (MoMask~\cite{guo2024momask} and BAMM~\cite{Pinyoanuntapong2024BAMMBA}). 
We also include a fine-tuning–based baseline, ESD~\cite{Gandikota2023ErasingCF}, for comparison. Drawing inspiration from discrete latent spaces now ubiquitous in motion generation, we introduce Latent Code Replacement (LCR), a novel, training-free approach that identifies violent codes in a VQ-VAE's codebook~\cite{denoord17vqvae} and substitutes them with safe alternatives, effectively erasing harmful behaviors while maintaining motion fidelity.

Our contributions are threefold:
\begin{itemize}
\item \textbf{Human Motion Unlearning.} We formulate the novel task of unlearning unsafe motion concepts from pretrained T2M models, contextualized through the challenge of violence prevention in motion synthesis.
\item \textbf{Violence Unlearning Benchmark.} We curate violence-filtered versions of HumanML3D and Motion-X and define metrics tailored to sequential motion unlearning, establishing a standardized framework to evaluate both suppression efficacy and motion quality.
\item \textbf{Exploration of Training-Free Methods and LCR.} We adapt UCE and RECE to T2M architectures, propose Latent Code Replacement as a new training-free paradigm leveraging discrete codebooks, and demonstrate that unlearning is indeed feasible; latent-code–based interventions offer the most promising performance, though substantial room for improvement remains.
\end{itemize}

% Related Works
\section{Related Works}
Human motion unlearning represents a convergence of machine unlearning and motion synthesis, two active research areas that have remained largely independent until now.

\paragraph{Machine Unlearning.} Machine unlearning aims to erase unwanted knowledge from generative models without compromising overall capabilities. This field has gained significant attention in image synthesis~\cite{sai2024machine,xu23survey1}, where approaches fall into three main categories: data removal methods that retrain models on filtered datasets~\cite{Guo2019CertifiedDR, chien2022certified}, fine-tuning approaches that adapt selected parameters~\cite{Gandikota2023ErasingCF, Lu2024mace, fan2024salun}, and training-free interventions that modify model behavior without additional training~\cite{gandikota2024uce, gong2024rece}.

Data removal quickly becomes impractical for large-scale corpora due to the sheer volume of samples and the prohibitive cost of manual annotation. Among fine-tuning methods, ESD~\cite{Gandikota2023ErasingCF} has been particularly influential. It updates a copy of the pretrained model in a contrastive fashion to penalize the generation of unsafe content. More recently, training-free techniques like UCE~\cite{gandikota2024uce} and RECE~\cite{gong2024rece} have emerged as SotA solutions. These methods rely on closed-form optimization to map undesirable concepts to predefined targets. Due to their efficiency and strong results, training-free methods now dominate the state of the art in image unlearning.
However, despite their success in image generation, these techniques have not yet been extended to sequence-based generative tasks, such as human motion synthesis, where the sequential and structured nature of the data introduces new challenges. In this work, we bridge this gap by adapting training-free unlearning methods to human motion synthesis and by proposing the first benchmark for motion unlearning.

\paragraph{Motion Synthesis.} 
Text-to-motion generation has made significant progress in recent years~\cite{petrovich22temos, tevet2023mdm, jiang2023motiongpt}, fueled by the availability of large-scale motion datasets and advances in deep generative modeling.
Modern motion synthesis approaches rely on either continuous~\cite{chen2023mld, sampieri24ladiff} or discrete latent representations~\cite{guo2024momask, cho2024discord, zhang2023t2mgpt}, with discrete latent spaces emerging as the dominant paradigm in SotA systems. Models like MoMask~\cite{guo2024momask} and BAMM~\cite{Pinyoanuntapong2024BAMMBA} use VQ-VAE codebooks~\cite{denoord17vqvae} to compress motion sequences into discrete tokens, enabling transformer architectures to efficiently capture long-range temporal dependencies and achieve superior scalability and generation quality. Despite the maturity of motion synthesis research, motion unlearning remains entirely unexplored. In this work, we introduce the first dedicated approach to motion unlearning, leveraging the discrete latent structure used in SotA models, and establish a benchmark for evaluating its effectiveness.

\paragraph{Ethics in Generative Models.} 
Recent work has highlighted the importance of safety considerations in generative models across modalities \cite{dixon18bias, bansal2022how}. Cultural differences in content perception and the challenges of defining ``harmful'' content universally have been noted in image generation contexts. Our work extends these considerations to the motion domain while acknowledging similar limitations.

% Motivation
\section{Human Motion Unlearning}

Human Motion Unlearning is the task of selectively removing specific types of motions from trained text-to-motion models while preserving generation quality on acceptable behaviors. Unlike image unlearning, HMU must address temporal dependencies where harmful patterns emerge across sequences rather than in static frames.

We formalize this task as follows.
Let $\mathcal{D} = {(t_i, m_i)}_{i=1}^{N}$ denote a dataset of text-motion pairs, where $t_i \in \mathcal{T}$ are textual prompts and $m_i \in \mathcal{M}$ are corresponding human motions. A T2M model $f_\theta$, parameterized by $\theta$, is trained to map text to motion:
\begin{equation}
    \label{eq:model:original}
    f_\theta \left(t \sim \mathcal{T}\right) = m \sim \mathcal{M}.
\end{equation}

We partition the dataset into a forget set $\mathcal{D}_f$ containing target concepts to be removed, and a retain set $\mathcal{D}_r = \mathcal{D} \setminus \mathcal{D}_f$ containing motions to be preserved. Motion unlearning seeks to reparameterize $\theta \rightarrow \theta'$ such that the updated model $f_{\theta'}$ no longer generates samples resembling the distribution of $\mathcal{D}_f$, while maintaining generation quality on $\mathcal{D}_r$.

\begin{figure}[!htb]
    \centering
    \includegraphics[width=0.47\textwidth]{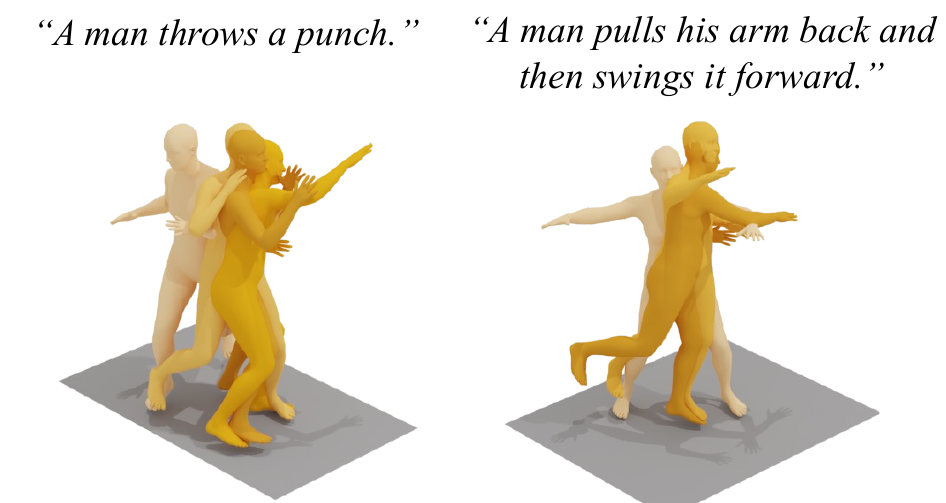}
    \caption{
    Explicit vs. implicit prompting of violent actions.
    }
    \label{fig:implicitprompt}
\end{figure}

An additional complication arises from implicit prompting, as shown in Figure \ref{fig:implicitprompt}, a phenomenon also found in images~\cite{yang24implicit}. Models may generate unwanted motions even when the prompt lacks harmful language. For example, while ``\textit{a man throws a punch}'' is explicitly violent, but a seemingly harmless alternative like ``\textit{a man pulls his arm back and then swings it forward}'' may produce the same output. These indirect descriptions effectively function as prompt injections, bypassing safety mechanisms by decomposing a violent act into a sequence of innocuous sub-motions.

\subsection{Benchmarking Violence Unlearning}

While HMU applies broadly, we focus on violent motions as our instantiation because: \textbf{i.} Text-to-motion (T2M) models can generate violent motions with serious social consequences, as motion synthesis datasets like HumanML3D~\cite{humanml3d} and Motion-X~\cite{motionx} contain violent behaviors, \textbf{ii.} violence offers rich complexity from atomic actions to compositional behaviors, and \textbf{iii.} as we will describe shortly, it provides measurable benchmarks for rigorous evaluation. This focus enables us to establish fundamental principles that generalize to other harmful content types.

\begin{figure}[!ht]

  \begin{subfigure}[t]{.23\textwidth}
    \centering
    \includegraphics[width=\linewidth]{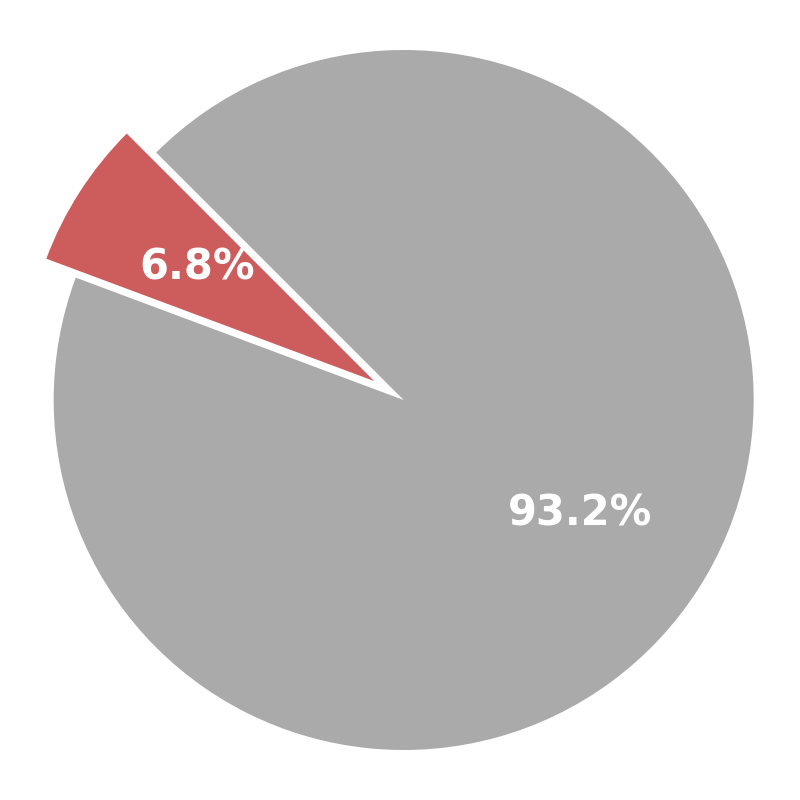}
    \caption{\scriptsize Violent behavior in HumanML3D.}
    \label{pie_t2m}
    
  \end{subfigure}
  \hfill
  \begin{subfigure}[t]{.23\textwidth}
    \centering
    \includegraphics[width=\linewidth]{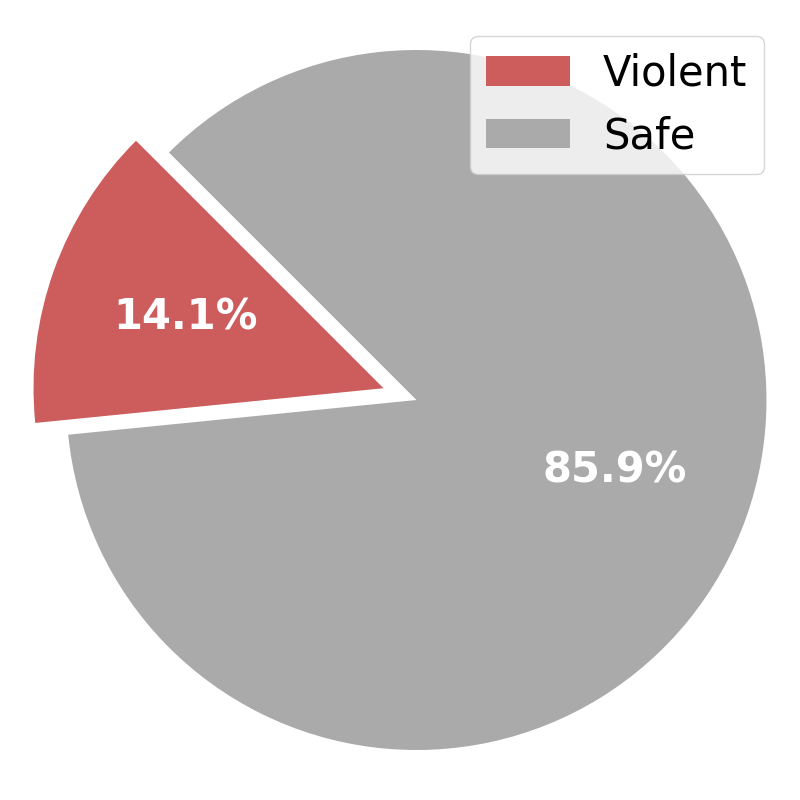}
    \caption{{\scriptsize Violent behavior in Motion-X.}}
    \label{pie_mx}
  \end{subfigure}
  \medskip
  \begin{subfigure}[t]{.23\textwidth}
    \centering
    \includegraphics[width=\linewidth]{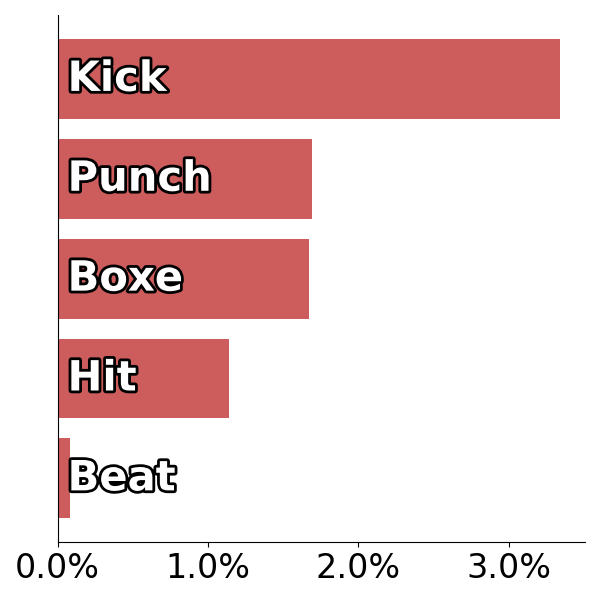}
    \caption{{\scriptsize Violent behavior in HumanML3D.}}
    \label{bar_t2m}
  \end{subfigure}
  \hfill
  \begin{subfigure}[t]{.23\textwidth}
    \centering
    \includegraphics[width=\linewidth]{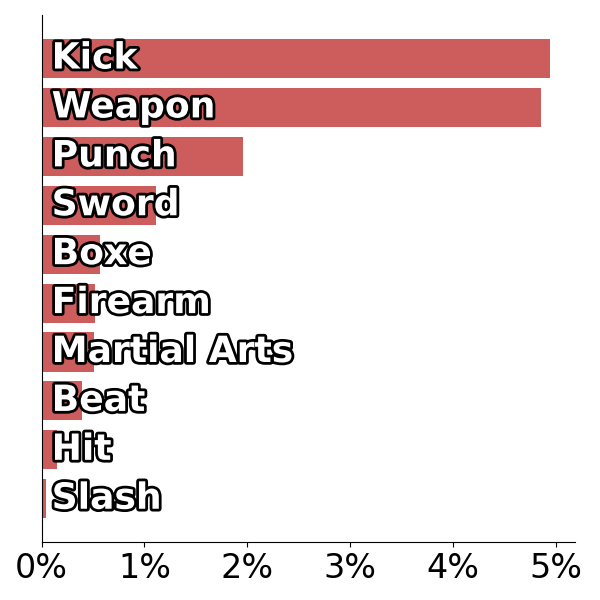}
    \caption{{\scriptsize Violent behavior in Motion-X.}}
    \label{bar_mx}
  \end{subfigure}
  \caption{Analysis of violent actions in HumanML3D and Motion-X. (Top row) Pie charts represent the proportion of harmful actions within each dataset. (Bottom row) Bar plots break down the occurrence of individual violent actions.}
\end{figure}

\paragraph{Datasets.}
We build our benchmark on two motion datasets: HumanML3D~\cite{humanml3d} is a widely used, large-scale dataset for text-to-motion (T2M) generation, containing approximately 15K motion-capture sequences paired with textual descriptions. Despite its popularity, it includes violent actions such as kicking, punching, and general combat behaviors. Our analysis shows that 7.7\% of the motions involve violence (Figure \ref{pie_t2m}), with kicking (3.4\%) and punching (1.7\%) being the most frequent (Figure \ref{bar_t2m}). While the remaining 92.3\% of samples are considered safe, the presence of violent behaviors poses risks when these are generated in unintended contexts.
Motion-X~\cite{motionx} has much larger scale and diversity. It contains around 81K sequences collected from a mix of sources, including videos scraped from the web. Notably, it has a much higher proportion of violent content, 14.9\% of its motions are classified as violent (Figure \ref{pie_mx}). These include kicking (4.9\%), use of weapons (5.0\%), and martial arts techniques like Kung Fu and Taekwondo (Figure \ref{bar_mx}). 
This highlights the growing risk of harmful behaviors being embedded as motion datasets scale, particularly when sourced from uncurated, real-world content.
Although Motion-X is less accessible and differs in format from HumanML3D, we process it into a compatible format and plan to release it alongside our benchmark.

For both datasets, we define a forget set $\mathcal{D}_f$ containing only violent motions, used to ensure the model forgets harmful behaviors, and a retain set $\mathcal{D}_r$, containing only safe motions, used to verify that the model preserves the quality of the safe content generation.
This filtering process is based on a predefined set of violent keywords ${w}$, which we curate manually and have already presented in Figures~\ref{bar_t2m} and~\ref{bar_mx}. A text-motion pair $(t, m)$ is assigned to $\mathcal{D}_f$ if at least one keyword $w_i$ appears in any prompt within $\mathcal{T}$.
Since motion is inherently sequential, violent sequences may contain both violent and non-violent sub-motions. Our benchmark assesses whether a model can suppress only violent segments while maintaining smooth transitions and general realism.

To evaluate robustness against implicit prompting, we use GPT-4 to rewrite explicit prompts into subtler, yet semantically equivalent formulations.

\subsection{Metrics}\label{sec:metrics}
We align with standard practices in text-to-motion and use FID to assess generation quality, Multimodality Distance (MM-Dist) and R-Precision to evaluate semantic alignment between text and motion, Diversity to measure variation across generations for different prompts, and Multimodality (MM) to measure variation among generations conditioned on the same prompt.
When evaluating performance on the forget set $\mathcal{D}_f$, we require a metric that ensures the model avoids generating violent motions. Traditional text-motion alignment metrics fall short here: a model that generates unrelated, random motions would achieve a better score than one that generates correct non-violent submotions while deliberately avoiding violent ones.
To address this, we modify MM-Dist by masking out violent words in the text prompts. 
``A person gives a \textit{kick}'' becomes ``A person gives a \textit{***}''. 
The resulting metric, called \MMnotox{}, evaluates whether the generated motion remains coherent with the non-violent parts of the prompt. On the retain set $\mathcal{D}_r$, \MMnotox{}  naturally reduces to standard MM-Dist, as there are no violent components to censor.

The following sections detail the state-of-the-art T2M models selected for evaluation and the unlearning methods adapted for our benchmark.

\begin{figure*}[!htb]
    \centering
    \includegraphics[width=0.85\textwidth]{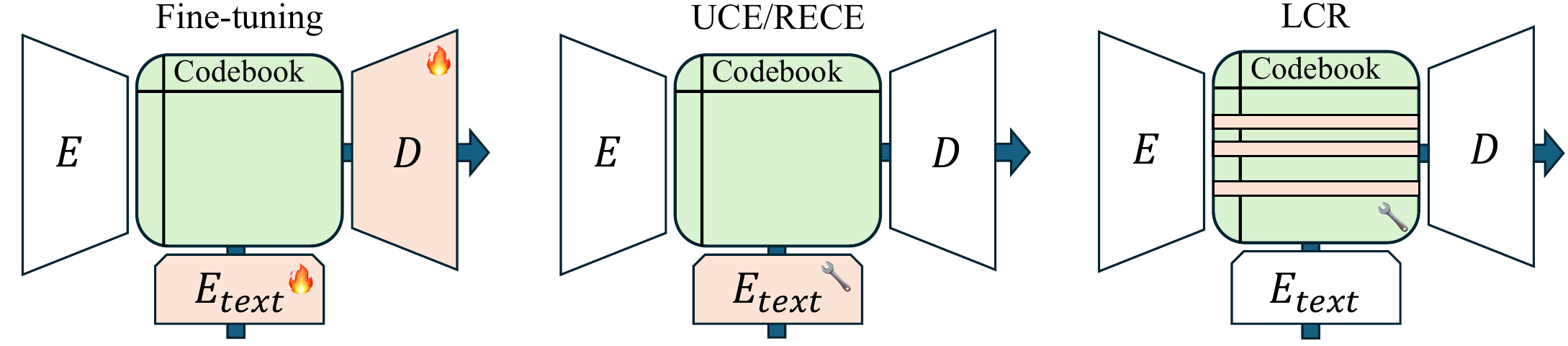}
    \caption{
    Illustration of motion unlearning approaches: (1) Fine-tuning modifies both the text encoder and motion decoder to remove violent actions, (2) UCE and RECE, as training-free methods, operate solely on the text encoder, (3) Our proposed LCR selectively updates only the affected codebook entries, ensuring targeted unlearning with minimal impact on overall synthesis quality.
    }
    \label{fig:pipeline}
\end{figure*}

% Text to Motion
\section{Text-to-Motion}
\label{sec:t2m}

Text-to-motion synthesis aims to generate 3D motion sequences from natural language descriptions of actions. Recent SotA models rely on discrete latent representations built using VQ-VAEs~\cite{denoord17vqvae}, which enable efficient token-based modeling of motion. Among them, MoMask~\cite{guo2024momask} and BAMM~\cite{Pinyoanuntapong2024BAMMBA} currently lead in performance.

\paragraph{Discrete Motion Representation.}
VQ-VAEs encode an input motion sequence $m \sim \mathcal{M}$ into a continuous latent sequence $Z = E(m) \in \mathbb{R}^{T \times d}$, where $T$ is the motion length and $d$ the embedding dimension. Each vector $z_t$ is quantized to its nearest codebook entry $c_{k_t}$ from a learned codebook $\mathcal{C} = \{c_n\}_{n=1}^N$ as:
\begin{equation}
    k_t = \arg\min_{j} \| z_t - c_j \|_2.
\end{equation}
This results in a discrete latent sequence $Z_q(m) = [k_1, \ldots, k_T]$, which is decoded by $D(\cdot)$ to reconstruct the original motion. Training optimizes three loss terms: a reconstruction loss, a commitment loss, and a codebook update loss~\cite{zhang2023t2mgpt, jiang2023motiongpt, guo2024momask}.

\paragraph{Text-Motion Alignment.}
Once the motion representation is learned, T2M models align it with textual descriptions to ensure semantic consistency. At inference, given a prompt $t$, a transformer generates a sequence of latent indices $Z_q$ autoregressively. This discrete sequence is then decoded using the VQ-VAE decoder to produce the final motion sequence.

% Unlearning Strategies
\section{Unlearning Strategies}
\label{sec:unlearnin_strat}

A growing trend in image machine unlearning is the transition from trainable to training-free methods. While the ideal solution would involve retraining a model from scratch on only safe data, this is typically infeasible due to the high computational cost, time demands, and the need for access to the complete training set. As a result, fine-tuning emerged as a practical alternative, enabling targeted forgetting without full retraining.
Recently, however, the state of the art has shifted towards training-free approaches, which achieve superior results with greater efficiency and speed. 
Most of these methods operate by reparametrizing cross-attention weights; we reimplement these methods in the human motion domain.
We also introduce a new perspective: instead of editing attention, we propose to redefine unlearning in the discrete latent space, characteristic of modern text-to-motion architectures.

\subsection{Attention-Driven Unlearning}
\label{sec:MUonT2M:RECE}

\paragraph{ESD.}
\cite{Gandikota2023ErasingCF} fine-tunes the model's weight matrices $W$ in a contrastive manner, pushing unwanted concepts $e_f$ away from a learned embedding space while pulling them toward a safe, predefined target $\bar{e}_r$. 
\begin{equation}
    \label{eq:esd}
    W^{new} \leftarrow  W^{old} e_f - \eta ( W^{old}\bar{e}_f - W^{old}\bar{e}_r ),
\end{equation}
 where $\eta$ is a guidance scale. Despite the rise of training-free methods, we include ESD and other trainable approaches in our benchmark for completeness.

\paragraph{UCE.}

\cite{gandikota2024uce} introduces a training-free solution by modifying the cross-attention projection matrix between text and latent representations. 
Given a set $F$ of concepts to forget, it aligns text embeddings $e_f \in F$ (e.g., ``punch someone'') to a target embedding $\bar{e}_r$ (e.g., the empty string), which has to be chosen a priori. UCE also retains other safe concepts in $R$.
\begin{equation}
    \label{eq:uce_open}
    \begin{aligned}
        \min_{W^{new}}  
        & \sum_{e_f \in F} || W^{new}e_f - W^{old}\bar{e}_r ||_2^2 \\
        & + \lambda_1 \sum_{e_r \in R} || W^{new}e_r - W^{old}e_r ||_2^2 \\
        & + \lambda_2 || W^{new} - W^{old} ||_F^2.
    \end{aligned}
\end{equation}
Hyperparameters $\lambda_1$ and $\lambda_2$ control the extent to which retained concepts are preserved. As shown in~\cite{gandikota2024uce}, a closed-form solution exists for Eq.~\ref{eq:uce_open}.  

\paragraph{RECE.} 
\cite{gong2024rece} extends UCE to improve robustness. UCE fails to fully forget a concept, as new embeddings $e'_f$ close to the forgotten one $e_f$ still trigger unwanted generations. RECE addresses this by iteratively searching for $e_f'$ near $e_f$ and applying UCE on them. This procedure is repeated for a predefined number of steps, progressively eliminating residual traces of the concept from the latent space.

\subsection{Latent Code Replacement in Human Motion}
\label{sec:MUonT2M:CP}

\begin{table*}[!ht]
\centering
\setlength{\tabcolsep}{1mm}
\fontsize{9}{11}\selectfont
\begin{tabularx}{\linewidth}{lc YYY YYYY}
\toprule
& & \multicolumn{3}{c}{Forget Set} & \multicolumn{4}{c}{Retain Set}     \\ \cmidrule(l{0.4cm}r{0.4cm}){3-5} \cmidrule(l{0.4cm}r{0.4cm}){6-9}
& Training-Free & FID $\rightarrow$ & \MMnotox $\downarrow$ & R@1 $\rightarrow$ & FID $\downarrow$   & MM-Dist $\downarrow$ & Diversity $\rightarrow$ & R@1 $\uparrow$  \\ 
\midrule
MoMask $\mathcal{D}_r$ & \xmark & 16.358$^{\pm.150}$ & 4.497$^{\pm.018}$ & 0.118$^{\pm.005}$ & 0.075$^{\pm.001}$ & 2.959$^{\pm.002}$ & 9.545$^{\pm.086}$ & 0.512$^{\pm.001}$ \\
\cline{3-9}
{\color[HTML]{\ColorTable} MoMask} &
{\color[HTML]{\ColorTable} --} &
${\color[HTML]{\ColorTable} 1.164^{\pm .048}}$ &
${\color[HTML]{\ColorTable} 5.593^{\pm .073}}$ &
${\color[HTML]{\ColorTable} 0.176^{\pm .006}}$ &
${\color[HTML]{\ColorTable} 0.041^{\pm .001}}$ &
${\color[HTML]{\ColorTable} 2.929^{\pm .002}}$ &
${\color[HTML]{\ColorTable} 9.629^{\pm .088}}$ &
${\color[HTML]{\ColorTable} 0.520^{\pm .001}}$ \\
MoMask \textit{FT} & \xmark & 2.295$^{\pm.065}$ & 5.002$^{\pm.016}$ & 0.150$^{\pm.006}$ & \underline{0.070}$^{\pm.001}$ & \underline{3.034}$^{\pm.003}$ & \underline{9.680}$^{\pm.111}$ & \underline{0.501}$^{\pm.002}$ \\
MoMask w/ ESD & \xmark & \underline{15.039}$^{\pm.122}$ & 6.397$^{\pm.033}$ & 0.071$^{\pm.005}$ & 30.679$^{\pm.110}$ & 7.378$^{\pm.011}$ & 6.673$^{\pm.044}$ & 0.165$^{\pm.001}$ \\
MoMask w/ UCE & \cmark & 11.860$^{\pm.154}$ & \textbf{4.626}$^{\pm.013}$ & \underline{0.135}$^{\pm.008}$ & 0.090$^{\pm.001}$ & 3.100$^{\pm.003}$ & 9.733$^{\pm.089}$ & 0.497$^{\pm.001}$ \\
MoMask w/ RECE & \cmark & 6.952$^{\pm.110}$ & 4.899$^{\pm.016}$ & 0.148$^{\pm.006}$ & 0.144$^{\pm.002}$ & 3.124$^{\pm.004}$ & 9.814$^{\pm.099}$ & 0.493$^{\pm.001}$ \\
MoMask w/ LCR & \cmark & \textbf{15.659}$^{\pm.128}$ & \underline{4.770}$^{\pm.019}$ &\textbf{0.124}$^{\pm.005}$ & \textbf{0.050}$^{\pm.001}$ & \textbf{2.986}$^{\pm.003}$ & \textbf{9.523}$^{\pm.084}$ & \textbf{0.508}$^{\pm.002}$ \\
\midrule\midrule
BAMM $\mathcal{D}_r$ & \xmark & 12.667$^{\pm.256}$ & 4.718$^{\pm.027}$ & 0.112$^{\pm.005}$ & 0.464$^{\pm.005}$ & 3.423$^{\pm.005}$ & 9.762$^{\pm.076}$ & 0.450$^{\pm.002}$ \\
\cline{3-9}
${\color[HTML]{\ColorTable} \text{BAMM}}$ &
${\color[HTML]{\ColorTable} \text{--}}$ &
${\color[HTML]{\ColorTable} 0.955^{\pm .055}}$ &
${\color[HTML]{\ColorTable} 4.995^{\pm .021}}$ &
${\color[HTML]{\ColorTable} 0.180^{\pm .008}}$ &
${\color[HTML]{\ColorTable} 0.181^{\pm .003}}$ &
${\color[HTML]{\ColorTable} 2.911^{\pm .003}}$ &
${\color[HTML]{\ColorTable} 9.731^{\pm .070}}$ &
${\color[HTML]{\ColorTable} 0.519^{\pm .001}}$
\\
BAMM \textit{FT} & \xmark & {1.081$^{\pm.009}$} & {5.000$^{\pm.049}$}  & {0.188$^{\pm.011}$} & 0.203$^{\pm.001}$ & {2.948}$^{\pm.000}$ & \underline{9.659}$^{\pm.004}$ & {0.516}$^{\pm.001}$ \\
BAMM w/ ESD & \xmark & 0.937$^{\pm.030}$ & 5.015$^{\pm.010}$ & 0.194$^{\pm.006}$ & {0.186$^{\pm.001}$} & \textbf{2.912$^{\pm.001}$} & {9.604$^{\pm.135}$} & \textbf{0.521$^{\pm.000}$} \\
BAMM w/ UCE & \cmark & 38.947$^{\pm.773}$ & 6.581$^{\pm.058}$  & \underline{0.088}$^{\pm.004}$ & 0.257$^{\pm.003}$ & 3.822$^{\pm.007}$ & 9.480$^{\pm.088}$ & 0.408$^{\pm.002}$ \\
BAMM w/ RECE & \cmark & \underline{1.428}$^{\pm.058}$ & \underline{4.965}$^{\pm.030}$ &0.178$^{\pm.006}$ & \textbf{0.170}$^{\pm.002}$ & 3.137$^{\pm.004}$ & 9.566$^{\pm.059}$ & 0.486$^{\pm.001}$ \\
BAMM w/ LCR & \cmark & \textbf{7.472}$^{\pm.525}$ & \textbf{4.647}$^{\pm.077}$ &\textbf{0.120}$^{\pm.005}$ & \underline{0.172}$^{\pm.001}$ & \textbf{2.912}$^{\pm.001}$ & \textbf{9.691}$^{\pm.166}$ & \underline{0.519}$^{\pm.004}$ \\
\bottomrule
\end{tabularx}

\caption{Results on HumanML3D dataset. We compare various unlearning strategies against the gold-standard ${\mathcal{D}_r}$, which is trained exclusively on violence-free motions. \textit{FT} denotes the results of fine-tuning the model on the a violence-free dataset. The original MoMask/BAMM models are in {\color[HTML]{\ColorTable} grey}. $\rightarrow$ means that the nearer to \textit{Method} $\mathcal{D}_r$ the better.}
\label{tab:humanml3d}
\end{table*}

Unlike previous approaches that modify attention parameters~\cite{gandikota2024uce, gong2024rece, Lu2024mace}, LCR operates directly on the codebook’s discrete latent space for precise representational  (see Figure~\ref{fig:pipeline}). Our method relies on two key assumptions: \textbf{i.} motion codes represent disentangled actions, and \textbf{ii.} violent motions are identifiable.
The first assumption follows from VQ-VAE's theory, where discretization inherently encourages concept disentanglement~\cite{tamkin2023codebook}. Empirical evidence supports this: removing violent codes preserves generation quality, while injecting them can compromise safe motions (see Appendix C%~\ref{sec:causal-intervention}
). This targeted latent intervention minimizes disruption to learned parameters while effectively eliminating violent patterns and maintaining high-quality generation for safe motions.

\paragraph{Detecting and Replacing Violent Motion Codes.}
Given a trained codebook $\mathcal{C}$, let $c \propto \mathcal{D}f$ indicate correlation between code $c \in \mathcal{C}$ and samples in $\mathcal{D}_f$. We want to identify the set of ``forget codes'' $\mathcal{C}_f = \{c \in \mathcal{C} \mid c \propto \mathcal{D}_f \}$.
For each code index $k$, we compute:
\begin{align}
    N_k(\mathcal{D}) & = \sum_{m \in \mathcal{D}} \mathds{1}\{k \in Z_q(m)\} \\
    s_k & = \frac{N_{k}(\mathcal{D}_f)}{N_{k}(\mathcal{D}_r)},
\end{align}
where $N_k(\mathcal{D})$ counts code index $k$’s occurrences in dataset $\mathcal{D}$, and $Z_q(m)$ represents codes activated by motion $m$. 
We select the top-$K$ codes with the highest $s_k$ values, which are those that frequently appear in violent motions but rarely in safe ones, to form the violent codebook $\mathcal{C}_f$.
Algorithm~\ref{alg:lcr} operates directly on the discrete latent space: each identified violent code is replaced with a randomly sampled safe code $\bar{c} \in \mathcal{C} \setminus \mathcal{C}_f$ plus noise $\varepsilon$ to ensure replacement uniqueness. This redirection prevents violent motion generation by substituting violent tokens with safe alternatives during the autoregressive generation process.

% input: algorithm
\begin{algorithm}[!b]
\caption{Latent Code Replacement (LCR)}
\label{alg:lcr}
\begin{algorithmic}[1]
\REQUIRE Trained codebook $\mathcal{C}$, forget dataset $\mathcal{D}_f$, retain dataset $\mathcal{D}_r$, number of codes to replace $K$.

\ENSURE Modified codebook with unlearned violent concepts

\STATE Compute $N_k(\mathcal{D}_f), N_k(\mathcal{D}_r) \;\;\;\forall k \in \mathcal{C}$

\STATE $s_k \leftarrow \frac{N_k(\mathcal{D}_f)}{N_k(\mathcal{D}_r)}$

\STATE $\mathcal{C}_f \leftarrow$ \texttt{Top-}$K( s_k)$

\STATE $\bar{c} \gets$ \texttt{sample}$( \mathcal{C} \setminus \mathcal{C}_f )$
\FOR{\textbf{each} $c_f$ in $\mathcal{C}_f$}
    \STATE $c_f \leftarrow \bar{c} + \varepsilon$
\ENDFOR

\RETURN $\mathcal{C}$ 
\end{algorithmic}
\end{algorithm}

\begin{table*}[!ht]
\centering
\setlength{\tabcolsep}{1mm}
\fontsize{9}{11}\selectfont
\begin{tabularx}{\linewidth}{lc YYY YYYY}
\toprule
& & \multicolumn{3}{c}{Forget Set} & \multicolumn{4}{c}{Retain Set}     \\ \cmidrule(l{0.4cm}r{0.4cm}){3-5} \cmidrule(l{0.4cm}r{0.4cm}){6-9}
& Training-Free & FID $\rightarrow$ & \MMnotox $\downarrow$ & R@1 $\rightarrow$ & FID $\downarrow$   & MM-Dist $\downarrow$ & Diversity $\rightarrow$ & R@1 $\uparrow$  \\ \midrule
\multicolumn{1}{l}{MoMask $\mathcal{D}_r$} & \xmark & 9.942$^{\pm.488}$ & 10.426$^{\pm.051}$ & 0.172$^{\pm.007}$ & 11.658$^{\pm.124}$ & 9.025$^{\pm.020}$ & 19.869$^{\pm.223}$ & 0.321$^{\pm.002}$ \\ \cline{3-9}
{\color[HTML]{\ColorTable} MoMask} &
{\color[HTML]{\ColorTable} -} &
${\color[HTML]{\ColorTable} 6.894^{\pm .338}}$ &
${\color[HTML]{\ColorTable} 9.291^{\pm .063}}$ &
${\color[HTML]{\ColorTable} 0.322^{\pm .011}}$ &
${\color[HTML]{\ColorTable} 3.697^{\pm .062}}$ &
${\color[HTML]{\ColorTable} 8.267^{\pm .021}}$ &
${\color[HTML]{\ColorTable} 19.343^{\pm .177}}$ &
${\color[HTML]{\ColorTable} 0.384^{\pm .003}}$
\\
\multicolumn{1}{l}{MoMask \textit{FT}} & \xmark & 33.433$^{\pm.675}$ & 12.838$^{\pm.045}$ & \textbf{0.184}$^{\pm.005}$ & 4.470$^{\pm.046}$ & \underline{8.992}$^{\pm.016}$ & 18.485$^{\pm.143}$ & \underline{0.337}$^{\pm.002}$ \\
\multicolumn{1}{l}{MoMask w/ ESD} & \xmark & 200.891$^{\pm1.421}$ & 17.977$^{\pm.029}$ & 0.029$^{\pm.004}$ & 172.559$^{\pm.535}$ & 19.001$^{\pm.014}$ & 6.512$^{\pm.076}$ & 0.032$^{\pm.001}$ \\
\multicolumn{1}{l}{MoMask w/ UCE} & \cmark & 53.451$^{\pm1.276}$ & 14.470$^{\pm.070}$ & \underline{0.148}$^{\pm.005}$ & 7.252$^{\pm.073}$ & 10.843$^{\pm.026}$ & 17.950$^{\pm.220}$ & 0.275$^{\pm.003}$ \\
\multicolumn{1}{l}{MoMask w/ RECE} & \cmark & \underline{13.415}$^{\pm.439}$ & \underline{11.205}$^{\pm.058}$ & {0.221}$^{\pm.008}$ & \underline{3.689}$^{\pm.056}$ & 9.142$^{\pm.019}$ & \underline{19.020}$^{\pm.182}$ & 0.332$^{\pm.002}$ \\
\multicolumn{1}{l}{MoMask w/ LCR} & \cmark & \textbf{7.078}$^{\pm.307}$ & \textbf{9.364}$^{\pm.066}$ & 0.317$^{\pm.009}$ &\textbf{3.658}$^{\pm.060}$ & \textbf{8.329}$^{\pm.017}$ & \textbf{19.344}$^{\pm.184}$ & \textbf{0.381}$^{\pm.003}$ \\
\bottomrule
\end{tabularx}
\caption{Results on Motion-X dataset. The same methods as in Table \ref{tab:humanml3d} are used. The original MoMask model is in {\color[HTML]{\ColorTable} grey}.}
\label{tab:motionx}
\end{table*}

% Results

\begin{figure}[ht]
    \centering
    \includegraphics[width=.48\textwidth]{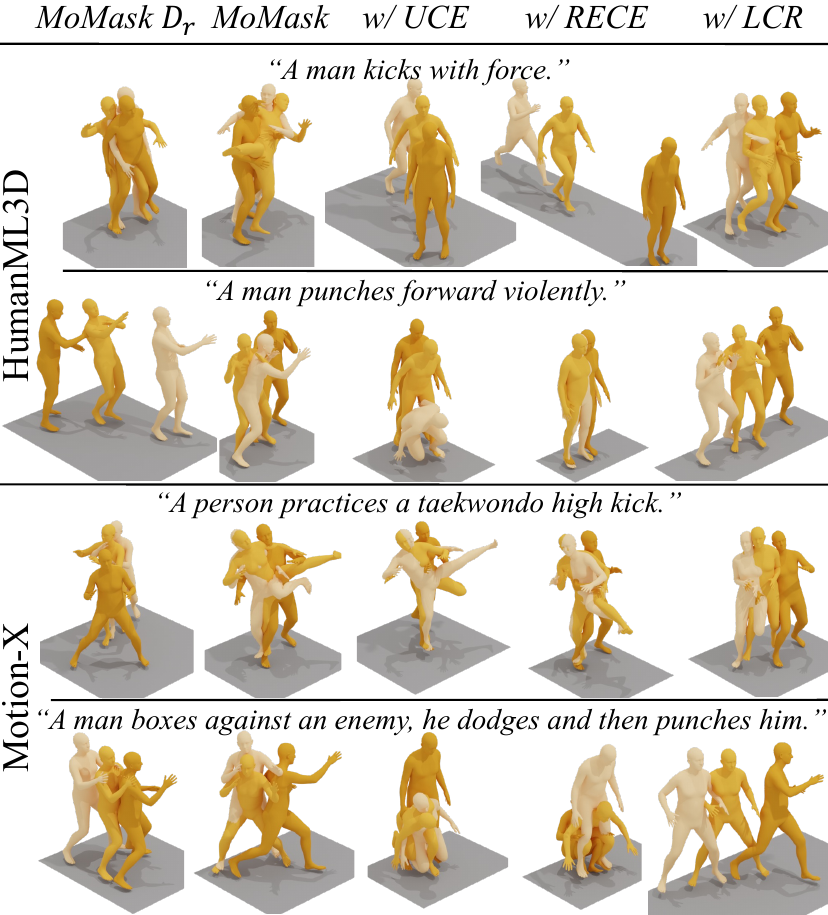}
    \caption{Qualitative comparison across datasets: HumanML3D and Motion-X samples demonstrating unlearning effectiveness. See videos on the project website.}
    \label{fig:comparison}
\end{figure}

\begin{figure}[ht]
    \centering
    \includegraphics[width=\linewidth]{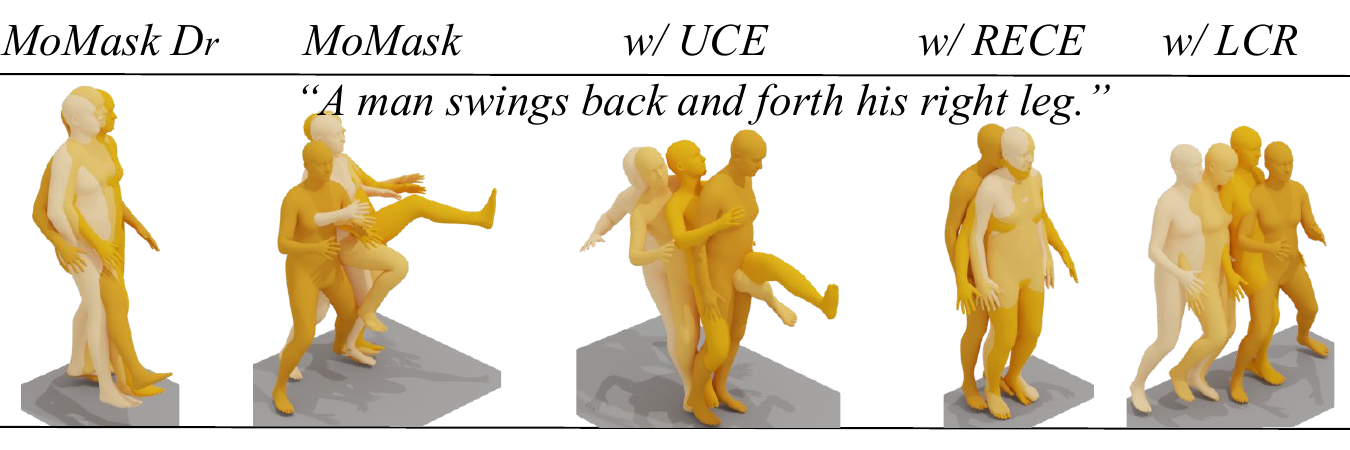}
    \caption{LCR successfully suppresses violent motions even when prompts are rephrased to avoid keywords.}
    \label{fig:implicit}
\end{figure}

\section{Results}
\label{sec:results}
In this section, we evaluate MoMask and BAMM on HumanML3D and Motion-X using all unlearning strategies from the previous section. Then, we provide qualitative results by comparing generated images from our method to those produced by other techniques. We establish an upper bound by training T2M models from scratch on the clean retain set $\mathcal{D}_r$, representing the ideal unlearning performance.

\paragraph{Forget Set Evaluation $\mathcal{D}_f$.} Our reference model \texttt{Method} $\mathcal{D}_r$ is trained from scratch on the retain set, and expectedly underperforms on violent actions since it never encountered them during training.
Unlearned models should closely mimic this upper bound behavior, effectively ``forgetting'' violent motions while maintaining baseline motion quality.
\paragraph{Retain Set Evaluation $\mathcal{D}_r$.} On non-violent motions, we apply standard T2M evaluation metrics~\cite{jiang2023motiongpt, sampieri24ladiff}. Since no violent prompts are involved, models should generate high-quality motions matching the textual descriptions without performance degradation. The \MMnotox{} reduces to the usual MM-Dist.

\paragraph{HumanML3D.} 
Table~\ref{tab:humanml3d} (top) evaluates MoMask with LCR unlearning. On the forget set, MoMask w/ LCR achieves the best FID and R@1 while maintaining strong \MMnotox{} scores, demonstrating effective violence reduction. On the retain set, LCR outperforms all baselines across all metrics, surpassing even MoMask $\mathcal{D}_r$ in FID.

Table~\ref{tab:humanml3d} (bottom) shows BAMM w/ LCR achieves optimal trade-off between motion quality and violence removal, with Retain FID of 0.172 and \MMnotox{} of 4.638. Notably, the low forget-set FID values for RECE, ESD, and \textit{FT} indicate failed unlearning of violent motions, whereas LCR's retain metrics closely match MoMask $\mathcal{D}_r$, confirming strong retention of non-forgotten data.

\paragraph{Motion-X.} 
Table~\ref{tab:motionx} presents Motion-X evaluation results. On the forget set, LCR scores 16.3\% lower than the second-best method on \MMnotox{}. On the retain set, LCR closely matches the original MoMask~\cite{guo2024momask}, preserving better than all the other alternative methods. BAMM unlearning results on Motion-X are provided in Appendix D%~\ref{sec:suppmat:full_tables}.
.

\paragraph{Qualitative Results.} Figure~\ref{fig:comparison} shows a qualitative comparison between LCR, UCE, and RECE. In HumanML3D, it is evident that the movements generated by LCR are similar to those of MoMask $\mathcal{D}_r$, where no violence is present. For the actions in Motion-X, it is noticeable that UCE and RECE either hint at violent actions or remain stationary without performing any. At the same time, LCR preserves the overall motion without introducing undesirable movements.

% Discussion
\section{Discussion}
\label{sec:discussion}
We analyze single-concept unlearning for both violent and non-violent actions, then evaluate robustness against implicit prompting. We focus on LCR due to its superior performance.

\paragraph{Single Concept Unlearning.} 
Table~\ref{tab:ablation:cp_toxicity} analyzes the impact of removing specific violent actions. We target the most common: \textit{kick} (3.4\% in HumanML3D, 4.9\% in Motion-X), and the least frequent (\textit{beat} in HumanML3D, \textit{stab} in Motion-X, both less than 0.2\%).\\
Removing \textit{kick} increases FID scores, suggesting reduced toxicity, while maintaining description consistency as MM-Safe remains stable. Also, removing \textit{beat} in HumanML3D is effective (high FID, low MM-Safe), whereas \textit{stab} in Motion-X proves harder to remove (low FID, low MM-Safe).
The forget set highlights how the impact varies by action, while the retain set confirms that safe actions remain unaffected, with performance aligning with the baseline.\\
To showcase broader applicability, we apply the same experiments to non-violent actions, and — in order to avoid any confusion — we provide the complete details, including the updated forget and retain set definitions, in Appendix A%\ref{sec:suppmat:nonviolent}
.

\begin{table}[!ht]
\centering
\setlength{\tabcolsep}{1mm}
\fontsize{9}{11}\selectfont
\begin{tabular}{lll cc cc}
\toprule
 &   &  & \multicolumn{2}{c}{Forget Set} & \multicolumn{2}{c}{Retain Set} \\ 
\cmidrule(l{0.2cm}r{0.2cm}){4-5} \cmidrule(l{0.2cm}r{0.2cm}){6-7}
 & & & FID $\rightarrow$   & \MMnotox $\downarrow$ & FID $\downarrow$ & MM-Dist $\downarrow$ \\ \midrule
 \multirow{4}*{\textbf{HML3D}} & {$\mathcal{D}_r$} & {\textit{viol.}}& {16.358} & {4.497} & {0.075} & {2.959} \\ \cline{3-7}
  &  \multirow{3}*{LCR} & \textit{viol.} &  15.659 & 4.769 & 0.050 & 2.986 \\
  & & \textit{kick} &  17.607 & 4.827  & 0.047 & 2.941 \\
  & &\textit{beat} & 13.309 & 4.557 &  0.201 & 3.133 \\ \midrule \midrule

 \multirow{4}*{\textbf{Motion-X}} & {$\mathcal{D}_r$} &\textit{viol.}& {9.942} & {10.426} & {11.658} & {9.025} \\ \cline{3-7}
& \multirow{3}*{LCR} & \textit{viol.} & 8.331 & 9.693 & 3.844 & 8.704 \\
 &  & \textit{kick} & 36.995 & 11.222 & 5.593& 9.039 \\
  & &  \textit{stab} & 24.315 & 7.312  & 3.624 & 8.314\\

\bottomrule 
\end{tabular}
\caption{Single concept unlearning results for MoMask on HumanML3D (\textit{top}) and Motion-X (\textit{bottom}).}
\label{tab:ablation:cp_toxicity}

\end{table}

\paragraph{Implicit Concept Unlearning.}
Naive filtering methods based on keyword detection are easily bypassed by rephrasing prompts to avoid banned terms. Since unlearning targets underlying concepts rather than specific words, robust methods must resist such circumvention. 
As shown in Figure~\ref{fig:implicit}, LCR withstands implicit prompting attacks suggesting that concept-level unlearning offers robustness than surface-level filtering.
Table 8 %~\ref{tab:reb:implicit}
in Appendix C compares explicit and implicit evaluations on a set of kicking actions. In the explicit case, our benchmark setup remains unchanged, and LCR outperforms MoMask consistently. In the implicit case, where the violent action is not named in the prompt, \MMnotox{} becomes equivalent to MM-Dist. 
Assuming the text embedding still captures the violent intent at the embedding level, a low MM-Dist would imply high alignment with a violent motion, something we want to avoid. Therefore, our goal is to push MM-Dist toward values similar to those of a model trained only on non-violent actions, indicating successful dissociation from violent intent.

% Conclusions
\section{Conclusions}
Text-to-motion models enable key applications but can reproduce violent behaviors from training data. We introduce Human Motion Unlearning to forget selected actions while preserving safe behaviors.
We present the first benchmark for motion unlearning with curated datasets and evaluation metrics, and propose Latent Code Replacement, a training-free method that edits motion codes to erase violent content without degrading quality.
LCR achieves the best safety-realism trade-off, laying the foundation for safe motion generation and broader unlearning research in generative temporal models.

% Ethical Statement
\section*{Ethical Statement}
By developing methods to selectively remove violent motions from generative models, our work aims to improve the safety of text-to-motion systems and reduce the computational cost of creating safer models. 
However, our approach carries potential risks. The capability to manipulate model outputs could be misused to introduce bias, censor legitimate content, or erase representation of specific demographics or cultural practices.
Additionally, our definition of ``violent'' motion reflects particular cultural assumptions that may not generalize across contexts, the same motion could be appropriate in martial arts training but harmful in other settings. While our method demonstrates feasibility, it does not provide complete safety guarantees and may fail to fully suppress targeted concepts or inadvertently degrade overall generation quality. 
We strongly recommend that text-to-motion models, with or without unlearning, should not be deployed in critical applications without multiple layers of safety controls and careful evaluation of their appropriateness for the specific use case.

% Acknowledgements
\section*{Acknowledgements}
We thank Luca Franco for the valuable insights and discussions.
We acknowledge support from Panasonic, the PNRR MUR project PE0000013-FAIR, and HPC resources provided by CINECA.

\bibliography{aaai2026}

\clearpage

\appendix
\section*{Appendices}
\section{Generalization to Non-Violent Motions}\label{sec:suppmat:nonviolent}

While our primary target application is on violent motions, due to their impact on society and presence in text-to-motion datasets, human motion unlearning is a general task that can target the removal of any action from the trained motion generation model.
In principle, it can be applied to any motion category deemed undesirable or inappropriate for a given application.
To demonstrate this, we evaluate our method on non-violent motion categories, showing that our approach generalizes well beyond the violence domain.
Quantitative results are reported in Table~\ref{tab:suppmat:non_violent}.

Specifically, we apply our LCR unlearning method to actions such as \textit{throw} (4.8\% of the dataset), \textit{jump} (8.5\%), and \textit{crawl} (1.8\%), selected for the generality and ambiguity of their associated prompts, which pose unique challenges for unlearning. 
On the forget set, \MMnotox{} achieves performance comparable to the reference model trained from scratch on their respective forget set $\mathcal{D}_r$. While the unlearning of \textit{throwing} is effective and stable, we observe a divergence in the FID of generated samples for \textit{jumping} and \textit{crawling}.

Despite these challenges, our method maintains strong performance on the retain set, matching that of the reference MoMask trained on $\mathcal{D}_r$, further reinforcing the robustness and generalizability of our approach beyond the violence domain.

\begin{table}[!ht]
\resizebox{.47\textwidth}{!}{
\begin{tabular}{ll cc cc}
\toprule
& & \multicolumn{2}{c}{Forget Set} & \multicolumn{2}{c}{Retain Set} \\ 
\cmidrule(l{0.4cm}r{0.4cm}){3-4} \cmidrule(l{0.4cm}r{0.4cm}){5-6}
& & FID $\rightarrow$ & \MMnotox $\downarrow$ & FID $\downarrow$ & MM-Dist $\downarrow$ \\ 
\midrule
\multirow{2}*{\textit{squat}} & {$\mathcal{D}_r$} & 4.382 & 5.080 & 0.072 & 2.959 \\
& LCR & 32.505 & 5.786 & 1.193 & 3.378 \\
\midrule
\multirow{2}*{\textit{throw}} & {$\mathcal{D}_r$} & 22.145 & 4.200 & 0.082 & 2.971 \\
& LCR & 21.693 & 5.464 & 0.171 & 3.091 \\
\midrule
\multirow{2}*{\textit{jump}} & {$\mathcal{D}_r$} & 21.458 & 4.616 & 0.136 & 3.055 \\
& LCR & 11.934 & 4.870 & 0.126 & 3.114 \\
\midrule
\multirow{2}*{\textit{crawl}} & {$\mathcal{D}_r$} & 12.280 & 3.839 & 0.068 & 3.006 \\
& LCR & 40.001 & 5.049 & 0.033 & 2.947 \\
\midrule
\multirow{2}*{\textit{zombie}} & {$\mathcal{D}_r$} & 25.623 & 5.451  & 0.042 & 2.932 \\
& LCR & 24.937 & 6.241 & 0.050 & 2.988 \\
\midrule \hline
\multirow{2}*{\textit{punch}} & {$\mathcal{D}_r$} & 17.653 & 4.147 & 0.095 & 2.977 \\
& LCR & 13.309 & 4.557 & 0.201 & 3.134 \\
\midrule
\multirow{2}*{\textit{kick}} & {$\mathcal{D}_r$} & 27.436 & 4.946 & 0.132 & 3.033 \\
& LCR & 17.607 & 4.827  & 0.047 & 2.941  \\
\bottomrule 
\end{tabular}}
\caption{Single concept unlearning results for MoMask on HumanML3D, LCR extends also to non-violent actions. 
}
\label{tab:suppmat:non_violent}
\end{table}

\section{LCR Ablation}\label{sec:lcr_ablation}

\paragraph{Varying the Number of Replaced Codes.}
Table~\ref{tab:cp_ablation} examines the impact of replacing different numbers of codes in LCR. Replacing 4 codes ($K=4$) improves the forget set, lowering FID compared to MoMask $\mathcal{D}_r$. However, the excessively low FID suggests the persistence of violent movements, whereas MoMask $\mathcal{D}_r$'s high FID reflects avoidance to generate them. Increasing replacements to $8 \leq K \leq 32$ reduces the FID gap and lowers \MMnotox{}, indicating better synthesis of non-violent movements. Beyond $K\geq 64$, performance declines, especially on the Retain Set.

\begin{table}[H]
\resizebox{\linewidth}{!}{
\begin{tabular}{lcccccc}
\toprule
 & & \multicolumn{4}{c}{Forget Set} \\ \cmidrule{3-6}
\textit{} & $K$ & FID $\rightarrow$ & \MMnotox{} $\downarrow$ & Diversity $\rightarrow$ & R@1 $\rightarrow$ \\ 
\midrule
MoMask $\mathcal{D}_r$ & -- & 13.412 & 4.735 & 5.909 & 0.132 \\ \hline
\multirow{5}{*}{LCR} 
 & 4  & 4.655  & 4.831 & \textbf{5.914} & 0.146 \\
 & 8  & 12.458 & 4.760 & 6.074          & 0.128 \\
 & 16 & \textbf{13.737} & 4.770 & 5.931 & \textbf{0.135} \\
 & 32 & 15.081 & \textbf{4.751} & 5.887 & 0.125 \\
 & 64 & 18.702 & 4.932 & 5.665 & 0.123 \\

\midrule
\toprule

 & & \multicolumn{4}{c}{Retain Set} \\ \cmidrule{3-6}
\textit{} & $K$ & FID $\downarrow$ & \MMnotox{} $\downarrow$ & Diversity $\rightarrow$ & R@1 $\uparrow$ \\ 
\midrule
MoMask $\mathcal{D}_r$ & -- & 0.047 & 2.960 & 9.636 & 0.514 \\
\hline
\multirow{5}{*}{LCR}
 & 4  & \textbf{0.041} & \textbf{2.931} & 9.612 & \textbf{0.519} \\
 & 8  & 0.045 & 2.941 & \textbf{9.635} & 0.517 \\
 & 16 & 0.047 & 2.960 & 9.652 & 0.515 \\
 & 32 & 0.051 & 2.986 & 9.506 & 0.507 \\
 & 64 & 0.106 & 3.045 & 9.399 & 0.496 \\

\bottomrule
\end{tabular}}
\caption{Ablation on number of codebook entries replaced (LCR) on HumanML3D.}
\label{tab:cp_ablation}
\end{table}

\paragraph{Discrete Compression Ratio}\label{sec:discrete-compression-ratio}
While the dimensionality of the codes in the codebook is set by the synthesis model, and not the unlearning method, we still ablate on the compression rate from continous to discrete.
In MoMask and BAMM, motion $m \in \mathbb{R}^{T \times D}$ is encoded as $c \in \mathbb{R}^{\frac{T}{2^s} \times 512}$ with stride $s = 2 $. We ablate $s$, retrain MoMask $\mathcal{D}_r$, and apply LCR. Tab.~\ref{tab:reb:codes} shows LCR stays close to MoMask $\mathcal{D}_r$, with $s=2$ performing best. 

\begin{table}[!ht]
\resizebox{.46\textwidth}{!}{
\begin{tabular}{l c c c c c}
\toprule
& & \multicolumn{2}{c}{Forget Set} & \multicolumn{2}{c}{Retain Set} \\ 
\cmidrule(lr){3-4} \cmidrule(lr){5-6}
\textit{} & $s$ & FID $\rightarrow$ & \MMnotox $\downarrow$ & FID $\downarrow$ & MM-Dist $\downarrow$ \\ 
\midrule
\multicolumn{1}{l}{MoMask $\mathcal{D}_r$} & 1 &
  $16.178^{\pm{.415}}$ & 
  $5.369^{\pm{.038}}$ & 
  $2.916^{\pm{.040}}$ & 
  $4.857^{\pm{.008}}$ \\
\multicolumn{1}{l}{MoMask w/ LCR} & 1 &
  $13.280^{\pm{.266}}$ & 
  $5.808^{\pm{.043}}$ & 
  $3.320^{\pm{.037}}$ & 
  $4.964^{\pm{.009}}$ \\
  \midrule
\multicolumn{1}{l}{MoMask $\mathcal{D}_r$} & 2 &
    16.358$^{\pm{.150}}$ & 
    4.497$^{\pm{.018}}$ & 
    0.075$^{\pm{.001}}$ & 
    2.959$^{\pm{.002}}$ \\
\multicolumn{1}{l}{MoMask w/ LCR} & 2 &
    15.659$^{\pm{.128}}$ &
    4.770$^{\pm{.019}}$ &
    0.050$^{\pm{.001}}$ &
    2.986$^{\pm{.003}}$ \\
\midrule
\multicolumn{1}{l}{MoMask $\mathcal{D}_r$} & 3 &
  $12.186^{\pm{.191}}$ & 
  $4.790^{\pm{.040}}$ & 
  $11.256^{\pm{.038}}$ & 
  $4.857^{\pm{.007}}$ \\
\multicolumn{1}{l}{MoMask w/ LCR} & 3 &
  $9.399^{\pm{.267}}$ & 
  $5.108^{\pm{.026}}$ 
  & $7.307^{\pm{.037}}$ 
  & $4.636^{\pm{.009}}$ \\
\bottomrule
\end{tabular}}
\caption{Ablation on HumanML3D of the compression rate.}
\label{tab:reb:codes}
\end{table}

\paragraph{Compositionality}\label{sec:composition}
A key question for LCR is whether it can effectively handle broad, heterogeneous categories like \textit{violence}, which encompasses multiple distinct motions (e.g., \textit{kick} and \textit{punch}). We investigate whether our code selection heuristic can precisely identify relevant codes across such varied motion sets, or whether decomposing the task into finer-grained concepts would improve performance.
We compare two strategies in Table \ref{tab:ablation:compositionality}. The \textit{direct approach} applies LCR to the entire \textit{violence} category at once with varying code replacements (16, 32, 64). The \textit{compositional approach} decomposes the task, applying LCR to \textit{kick} and \textit{punch} individually before combining interventions (e.g., 16 codes for \textit{kick} and 8 for \textit{punch}).
The results show that the compositional approach (LCR \textit{composed} 16 \& 8) yields only marginal gains in forget performance over the direct approach, while retain performance remains identical. This demonstrates that our code selection heuristic is robust and precise enough to identify relevant codes for diverse motions within a broad category. The added complexity of compositional decomposition is unnecessary, LCR effectively unlearns heterogeneous categories directly.

\begin{table}[!ht]
\resizebox{\linewidth}{!}{
\begin{tabular}{lcc cc cc}
\toprule
& \multicolumn{2}{c}{\# codes} & \multicolumn{2}{c}{Forget Set} & \multicolumn{2}{c}{Retain Set} \\
\cmidrule(l{0.3cm}r{0.4cm}){2-3} \cmidrule(l{0.3cm}r{0.4cm}){4-5} \cmidrule(l{0.3cm}r{0.4cm}){6-7}
& \textit{kick} & \textit{punch} & FID $\rightarrow$& \MMnotox $\downarrow$ & FID $\downarrow$ & MM-Dist $\downarrow$ \\ \midrule
MoMask $\mathcal{D}_r$ & -- & -- & 16.358 & 4.497 & 0.075 & 2.959 \\ \hline
\multirow{3}*{LCR}& \multicolumn{2}{c}{16} & 13.737 & 4.770 & 0.047 & 2.960 \\
 & \multicolumn{2}{c}{32} & 15.081 & 4.751 & 0.051 & 2.986 \\
& \multicolumn{2}{c}{64} & 18.702 & 4.932 & 0.106 & 3.045 \\
\hline

\multirow{2}*{LCR \textit{composed}} & 16 & 8 & 13.412 & 4.735 & 0.047 & 2.960\\
& 32 & 16 & 15.659 & 4.770 & 0.050 & 2.986\\
\bottomrule
\end{tabular}}
\caption{LCR unlearning results for the \textit{violence} category. \textbf{Top:} Performance with increasing numbers of code replacements applied to the entire \textit{violence} category. \textbf{Bottom:} Compositional results when combining LCR applications from individual concepts (\textit{punch} and \textit{kick}) within the \textit{violence} category.}
\label{tab:ablation:compositionality}
\end{table}

\section{Extra Qualitative Results}

\begin{figure}[!ht]
    \centering
    \includegraphics[width=0.47\textwidth]{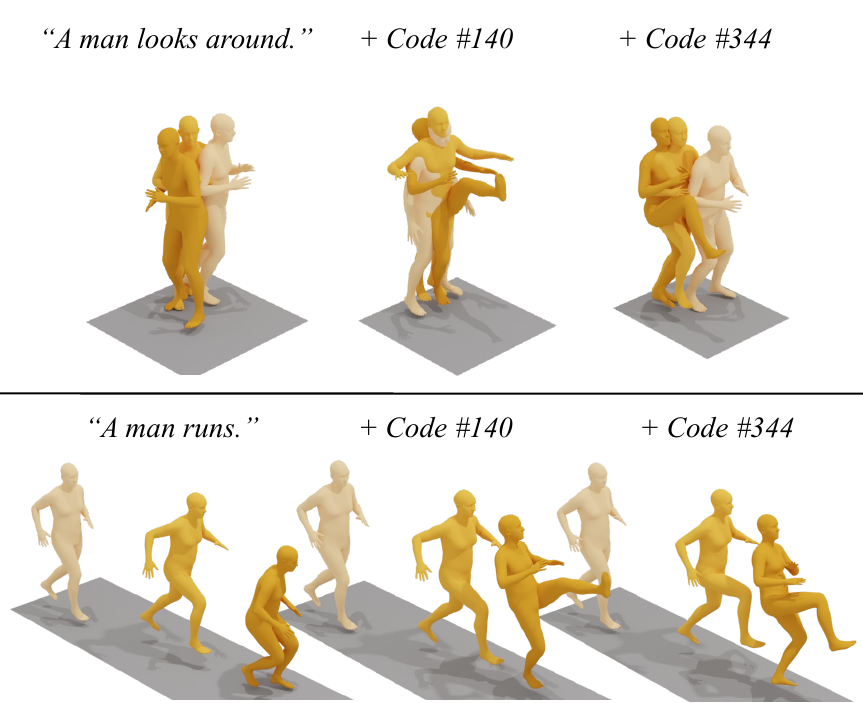}
    \caption{Examples of causal intervention. Injecting the violent code into the model at inference, causes the model to produce violent motions.}
    \label{fig:inject}
\end{figure}

\paragraph{Causal Intervention.}
\label{sec:causal-intervention}
Our LCR approach is based on the intuition that codes in a discrete space are disentangled, and can represent specific actions. For this reason, we perform unlearning by replacing codes associated with violence.
To further support this, we show violence injection on human motion. First, we identify codes associated to a toxic motion, like kicking. Then, given a safe prompt, we inject toxic codes in the latent sequence $Z$.
In Figure \ref{fig:inject} we show that violence injection: we can see that even though the given prompts are safe, adding specific codes allows the model to perform kicks. Specifically, code 140 represents a high kick with the left leg, while code 344 represents a lower kick with the right leg.

\paragraph{Robustness to Implicit Prompting Attacks.}
Section~\ref{sec:discussion} of the main paper introduces an ablation study on implicit concept unlearning and demonstrates how prompt injection attacks can circumvent safety measures. Here in the appendix, we provide additional quantitative results in Table~\ref{tab:reb:implicit}, contrasting model performance under explicit versus implicit prompting conditions.
To generate implicit prompts, we adopt the following procedure: we first identify prompts associated with violent actions, then employ GPT-4 to rephrase them such that the violence is conveyed indirectly rather than through explicit terminology. For instance, a kicking motion is reformulated as ``A person swings their right leg back and forth'', as illustrated in Figure~\ref{fig:implicit}. 
This approach tests whether the model has truly unlearned the underlying concept or merely learned to filter specific keywords. We encourage readers to visit the project website for video demonstrations of these implicit prompting scenarios.

\begin{table}[!ht]
\centering
\setlength{\tabcolsep}{1mm}
\fontsize{9}{11}\selectfont
\begin{tabular}{lcccc}
\toprule
& \multicolumn{2}{c}{Explicit Violence} & \multicolumn{2}{c}{Implicit Violence} \\ 
\cmidrule(lr){2-3} \cmidrule(lr){4-5}
& FID $\rightarrow$ & \MMnotox{} $\downarrow$ & FID $\rightarrow$ & MM-Dist $\rightarrow$ \\ 
\midrule
\multicolumn{1}{l}{MoMask $\mathcal{D}_r$} &
  22.523 &
  4.608 &
  21.738 &
  5.026 \\
\midrule

\multicolumn{1}{l}{{\color[HTML]{\ColorTable} MoMask}} &
{\color[HTML]{\ColorTable} 1.032} &
{\color[HTML]{\ColorTable} 5.194} &
{\color[HTML]{\ColorTable} 7.706} &
{\color[HTML]{\ColorTable} 4.161} \\
\multicolumn{1}{l}{MoMask w/ LCR} &
    {18.652} &
    {4.868} &
    {25.782} &
    {5.351} \\
\bottomrule

\end{tabular}
\caption{Explicit v. implicit on the set of \textit{kick} prompts.}
\label{tab:reb:implicit}
\end{table}
\section{Full Tables} \label{sec:suppmat:full_tables}
\begin{table*} [!htp]
\resizebox{1.\textwidth}{!}{
\begin{tabular}{lc ccccccc}
\toprule\textit{} &
  \multicolumn{7}{c}{Forget Set} \\ \cmidrule(l{.4cm}r{.4cm}){3-9}
\textit{} & Training-free &
  FID $\rightarrow$ &
  \MMnotox{} $\downarrow$ &
  Diversity $\rightarrow$ &
  MultiModality $\rightarrow$ &
  R@1 $\rightarrow$ &
  R@2 $\rightarrow$ &
  R@3 $\rightarrow$ \\ \midrule
\multicolumn{1}{l}{MoMask $\mathcal{D}_r$} & \xmark & 16.358$^{\pm{.150}}$ & 4.497$^{\pm{.018}}$ & 6.955$^{\pm{.054}}$ & 2.078$^{\pm{.000}}$ & 0.118$^{\pm{.005}}$ & 0.208$^{\pm{.006}}$ & 0.278$^{\pm{.007}}$ \\ \hline
\multicolumn{1}{l}{\color[HTML]{\ColorTable} MoMask} & \color[HTML]{\ColorTable} - & \color[HTML]{\ColorTable} 1.164$^{\pm{.048}}$ & \color[HTML]{\ColorTable} 5.107$^{\pm{.015}}$ & \color[HTML]{\ColorTable} 5.593$^{\pm{.073}}$ & \color[HTML]{\ColorTable} 1.616$^{\pm{.000}}$ & \color[HTML]{\ColorTable} 0.176$^{\pm{.006}}$ & \color[HTML]{\ColorTable} 0.293$^{\pm{.006}}$ & \color[HTML]{\ColorTable} 0.382$^{\pm{.008}}$ \\
\multicolumn{1}{l}{MoMask \textit{FT}} & \xmark & 2.295$^{\pm{.065}}$ & 5.002$^{\pm{.016}}$ & 5.915$^{\pm{.055}}$ & 1.767$^{\pm{.000}}$ & 0.150$^{\pm{.006}}$ & 0.255$^{\pm{.006}}$ & 0.346$^{\pm{.007}}$ \\
\multicolumn{1}{l}{MoMask w/ ESD} & \xmark & \underline{15.039}$^{\pm{.122}}$ & 6.397$^{\pm{.033}}$ & 5.100$^{\pm{.067}}$ & \underline{1.947}$^{\pm{.000}}$ & 0.071$^{\pm{.005}}$ & 0.136$^{\pm{.010}}$ & 0.181$^{\pm{.008}}$ \\
\multicolumn{1}{l}{MoMask w/ UCE} & \cmark & 11.860$^{\pm{.154}}$ & \textbf{4.626$^{\pm{.013}}$} & \textbf{7.144$^{\pm{.080}}$} & \textbf{2.069$^{\pm{.000}}$} & \underline{0.135}$^{\pm{.008}}$ & \underline{0.221}$^{\pm{.007}}$ & \textbf{0.289$^{\pm{.005}}$} \\
\multicolumn{1}{l}{MoMask w/ RECE} & \cmark & 6.952$^{\pm{.110}}$ & 4.899$^{\pm{.016}}$ & \underline{6.548}$^{\pm{0.048}}$ & 1.900$^{\pm{.000}}$ & 0.148$^{\pm{.006}}$ & 0.245$^{\pm{.007}}$ & 0.321$^{\pm{.008}}$ \\
\multicolumn{1}{l}{MoMask w/ LCR} & \cmark & \textbf{15.659$^{\pm{.165}}$} & \underline{4.770}$^{\pm{.023}}$ & 5.997$^{\pm{.065}}$ & 2.219$^{\pm{.000}}$ & \textbf{0.125$^{\pm{.005}}$} & \textbf{0.215$^{\pm{.006}}$} & \underline{0.291}$^{\pm{.006}}$ \\ \midrule \midrule

\multicolumn{1}{l}{BAMM $\mathcal{D}_r$} & \xmark & 12.667$^{\pm{.256}}$ & 4.718$^{\pm{.027}}$ & 7.347$^{\pm{.070}}$ & 6.544$^{\pm{.000}}$ & 0.112$^{\pm{.005}}$ & 0.188$^{\pm{.005}}$ & 0.251$^{\pm{.006}}$ \\ \hline
\multicolumn{1}{l}{\color[HTML]{\ColorTable}BAMM} & \color[HTML]{\ColorTable} - & \color[HTML]{\ColorTable} 0.955$^{\pm{.055}}$ & \color[HTML]{\ColorTable} 4.995$^{\pm{.021}}$ & \color[HTML]{\ColorTable} 5.600$^{\pm{.069}}$ & \color[HTML]{\ColorTable} 4.855$^{\pm{.000}}$ & \color[HTML]{\ColorTable} 0.180$^{\pm{.008}}$ & \color[HTML]{\ColorTable} 0.297$^{\pm{.006}}$ & \color[HTML]{\ColorTable} 0.392$^{\pm{.008}}$ \\
\multicolumn{1}{l}{ BAMM \textit{FT}} & \xmark & 1.081$^{\pm{.009}}$ & 5.000$^{\pm{.049}}$ & 5.725$^{\pm{.285}}$ & 5.095$^{\pm{.000}}$ & 0.188$^{\pm{.011}}$ & 0.315$^{\pm{.013}}$ & 0.409$^{\pm{.029}}$ \\
\multicolumn{1}{l}{ BAMM w/ ESD} & \xmark & 0.937$^{\pm{.030}}$ & 5.015$^{\pm{.010}}$ & 5.569$^{\pm{.051}}$ & 4.847$^{\pm{.000}}$ & 0.194$^{\pm{.006}}$ & 0.312$^{\pm{.008}}$ & 0.402$^{\pm{.013}}$ \\
\multicolumn{1}{l}{ BAMM w/ UCE} & \cmark & 38.947$^{\pm{.773}}$ & 6.581$^{\pm{.058}}$ & \underline{8.807}$^{\pm{.084}}$ & \underline{7.750}$^{\pm{.000}}$ & \underline{0.088}$^{\pm{.004}}$ & \underline{0.157}$^{\pm{.004}}$ & \underline{0.207}$^{\pm{.005}}$ \\
\multicolumn{1}{l}{ BAMM w/ RECE} & \cmark & \underline{1.428}$^{\pm{.058}}$ & \underline{4.965}$^{\pm{.030}}$ & 5.623$^{\pm{.062}}$ & 4.910$^{\pm{.000}}$ & 0.178$^{\pm{.006}}$ & 0.291$^{\pm{.007}}$ & 0.373$^{\pm{.008}}$ \\
\multicolumn{1}{l}{ BAMM w/ LCR} & \cmark & \textbf{7.742$^{\pm{.005}}$} & \textbf{4.647$^{\pm{.008}}$} & \textbf{6.198$^{\pm{.070}}$} & \textbf{5.394$^{\pm{.087}}$} & \textbf{0.120$^{\pm{.005}}$} & \textbf{0.213$^{\pm{.027}}$} & \textbf{0.290$^{\pm{.012}}$} \\
\bottomrule
\end{tabular}}
\caption{Forget set full comparison on HumanML3D.}
\label{tab:suppmat:t2m_forget}
\end{table*}
\begin{table*}
\resizebox{1.\textwidth}{!}{
\begin{tabular}{lc ccccccc}
\toprule
\textit{} &
  \multicolumn{7}{c}{Retain Set} \\ 
  \cmidrule(l{0.4cm}r{0.4cm}){3-9}
\textit{} & Training-free &
  FID $\downarrow$ &
  MM-Dist $\downarrow$ &
  Diversity $\rightarrow$ &
  MultiModality $\uparrow$ &
  \multicolumn{1}{c}{R@1 $\uparrow$} &
  R@2 $\uparrow$ &
  R@3 $\uparrow$ \\ \midrule
\multicolumn{1}{l}{MoMask $\mathcal{D}_r$} & \xmark & 
  0.075$^{\pm.001}$ & 
  2.959$^{\pm.002}$ & 
  9.545$^{\pm.086}$ & 
  1.194$^{\pm.047}$ & 
  0.512$^{\pm.001}$ & 
  0.706$^{\pm.001}$ & 
  0.801$^{\pm.001}$ \\ \midrule
\multicolumn{1}{l}{{\color[HTML]{\ColorTable} MoMask}} & \color[HTML]{\ColorTable} - & 
  \color[HTML]{\ColorTable} 0.041$^{\pm.001}$ & 
  \color[HTML]{\ColorTable} 2.929$^{\pm.002}$ & 
  \color[HTML]{\ColorTable} 9.629$^{\pm.088}$ & 
  \color[HTML]{\ColorTable} 1.218$^{\pm.046}$ & 
  \color[HTML]{\ColorTable} 0.520$^{\pm.001}$ & 
  \color[HTML]{\ColorTable} 0.713$^{\pm.001}$ & 
  \color[HTML]{\ColorTable} 0.807$^{\pm.001}$\\
\multicolumn{1}{l}{MoMask \textit{FT}} & \xmark & 
  \underline{0.070}$^{\pm.001}$ & 
  \underline{3.034}$^{\pm.003}$ & 
  \underline{9.680}$^{\pm.111}$ & 
  \underline{1.237}$^{\pm.044}$ & 
  \underline{0.501}$^{\pm.002}$ & 
  \underline{0.695}$^{\pm.001}$ & 
  \underline{0.794}$^{\pm.001}$ \\
\multicolumn{1}{l}{MoMask w/ ESD} & \xmark & 
  30.679$^{\pm.110}$ & 
  7.378$^{\pm.011}$ & 
  6.673$^{\pm.044}$ & 
  \textbf{2.211}$^{\pm.075}$ & 
  0.165$^{\pm.001}$ & 
  0.260$^{\pm.001}$ & 
  0.328$^{\pm.000}$ \\
\multicolumn{1}{l}{MoMask w/ UCE} & \cmark & 
  0.090$^{\pm.001}$ & 
  3.100$^{\pm.003}$ & 
  9.733$^{\pm.089}$ & 
  1.181$^{\pm.044}$ & 
  0.497$^{\pm.001}$ & 
  0.685$^{\pm.001}$ & 
  0.780$^{\pm.001}$ \\
\multicolumn{1}{l}{MoMask w/ RECE} & \cmark & 
  0.144$^{\pm.002}$ & 
  3.124$^{\pm.004}$ & 
  9.814$^{\pm.099}$ & 
  1.198$^{\pm.043}$ & 
  0.493$^{\pm.001}$ & 
  0.683$^{\pm.002}$ & 
  0.780$^{\pm.001}$ \\
\multicolumn{1}{l}{MoMask w/ LCR} & \cmark & 
  \textbf{0.050}$^{\pm.001}$ & 
  \textbf{2.986}$^{\pm.003}$ & 
  \textbf{9.523}$^{\pm.084}$ & 
  1.226$^{\pm.053}$ & 
  \textbf{0.508}$^{\pm.002}$ & 
  \textbf{0.700}$^{\pm.001}$ & 
  \textbf{0.797}$^{\pm.001}$ \\ \midrule \midrule
\multicolumn{1}{l}{BAMM $\mathcal{D}_r$} & \xmark & 
  0.464$^{\pm.005}$ & 
  3.423$^{\pm.005}$ & 
  9.762$^{\pm.076}$ & 
  8.436$^{\pm.198}$ & 
  0.450$^{\pm.002}$ & 
  0.640$^{\pm.001}$ & 
  0.743$^{\pm.001}$ \\ \hline
\multicolumn{1}{l}{{\color[HTML]{\ColorTable} BAMM}} & \color[HTML]{\ColorTable} - & 
  \color[HTML]{\ColorTable} 0.181$^{\pm.003}$ & 
  \color[HTML]{\ColorTable} 2.911$^{\pm.003}$ & 
  \color[HTML]{\ColorTable} 9.731$^{\pm.070}$ & 
  \color[HTML]{\ColorTable} 8.105$^{\pm.252}$ & 
  \color[HTML]{\ColorTable} 0.519$^{\pm.001}$ & 
  \color[HTML]{\ColorTable} 0.714$^{\pm.001}$ & 
  \color[HTML]{\ColorTable} 0.809$^{\pm.001}$\\
\multicolumn{1}{l}{BAMM \textit{FT}} & \xmark & 
  0.203$^{\pm.001}$ & 
  2.948$^{\pm.000}$ & 
  \underline{9.659}$^{\pm.004}$ & 
  \underline{8.368}$^{\pm.147}$ & 
  0.516$^{\pm.001}$ & 
  0.709$^{\pm.001}$ & 
  0.805$^{\pm.002}$ \\
\multicolumn{1}{l}{BAMM w/ ESD} & \xmark & 
  0.186$^{\pm.001}$ & 
  \textbf{2.912}$^{\pm.001}$ & 
  9.604$^{\pm.135}$ & 
  8.067$^{\pm.747}$ & 
  \textbf{0.521}$^{\pm.000}$ & 
  \underline{0.714}$^{\pm.001}$ & 
  \underline{0.810}$^{\pm.004}$ \\
\multicolumn{1}{l}{BAMM w/ UCE} & \cmark & 
  0.257$^{\pm.003}$ & 
  3.822$^{\pm.007}$ & 
  9.480$^{\pm.088}$ & 
  8.011$^{\pm.249}$ & 
  0.408$^{\pm.002}$ & 
  0.582$^{\pm.001}$ & 
  0.680$^{\pm.001}$ \\
\multicolumn{1}{l}{BAMM w/ RECE} & \cmark & 
  \textbf{0.170}$^{\pm.002}$ & 
  3.137$^{\pm.004}$ & 
  9.566$^{\pm.059}$ & 
  7.913$^{\pm.154}$ & 
  0.486$^{\pm.001}$ & 
  0.678$^{\pm.002}$ & 
  0.777$^{\pm.002}$ \\
\multicolumn{1}{l}{BAMM w/ LCR} & \cmark & 
  \underline{0.172}$^{\pm.001}$ & 
  \textbf{2.912}$^{\pm.001}$ & 
  \textbf{9.691}$^{\pm.166}$ & 
  \textbf{8.552}$^{\pm.097}$ & 
  \underline{0.519}$^{\pm.004}$ & 
  \textbf{0.716}$^{\pm.004}$ & 
  \textbf{0.812}$^{\pm.003}$ \\ \bottomrule
\end{tabular}}
\caption{Retain set full comparison HumanML3D.}
\label{tab:suppmat:t2m_retain}
\end{table*}

\begin{table*}
\resizebox{1.\textwidth}{!}{
\begin{tabular}{lc ccccccc}
\toprule
\textit{} &
  \multicolumn{7}{c}{Forget Set} \\   \cmidrule(l{0.4cm}r{0.4cm}){3-9}

\textit{} & Training-free & 
  FID $\rightarrow$ & 
  \MMnotox{} $\downarrow$ &
  Diversity $\rightarrow$ &
  MultiModality $\rightarrow$ &
  R@1 $\rightarrow$ &
  R@2 $\rightarrow$ &
  R@3 $\rightarrow$ \\ \midrule
\multicolumn{1}{l}{MoMask $\mathcal{D}_r$} & \xmark & 9.942$^{\pm{.488}}$ & 10.426$^{\pm{.051}}$ & 17.190$^{\pm{.117}}$ & 4.977$^{\pm{.000}}$ & 0.172$^{\pm{.007}}$ & 0.286$^{\pm{.008}}$ & 0.366$^{\pm{.008}}$ \\ \hline
\multicolumn{1}{l}{\color[HTML]{\ColorTable} MoMask} & \color[HTML]{\ColorTable} -- & \color[HTML]{\ColorTable} 6.894$^{\pm{.338}}$ & \color[HTML]{\ColorTable} 9.291$^{\pm{.063}}$ & \color[HTML]{\ColorTable} 17.113$^{\pm{.156}}$ & \color[HTML]{\ColorTable} 3.702$^{\pm{.000}}$ & \color[HTML]{\ColorTable} 0.322$^{\pm{.011}}$ & \color[HTML]{\ColorTable} 0.488$^{\pm{.012}}$ & \color[HTML]{\ColorTable} 0.591$^{\pm{.012}}$ \\

\multicolumn{1}{l}{MoMask \textit{FT}} & \xmark & 33.433$^{\pm.675}$ & {12.838}$^{\pm.045}$ & 
16.723$^{\pm.189}$ &
3.826$^{\pm.000}$ & 
\textbf{0.184}$^{\pm.005}$ & 
\textbf{0.293}$^{\pm.008}$ & 
\textbf{0.328}$^{\pm.012}$
\\

\multicolumn{1}{l}{MoMask w/ ESD} & \xmark & 200.891$^{\pm{1.421}}$ & 17.977$^{\pm{.029}}$ & 6.261$^{\pm{.074}}$ & 2.539$^{\pm{.000}}$ & 0.029$^{\pm{.004}}$ & 0.059$^{\pm{.006}}$ & 0.089$^{\pm{.007}}$ \\
\multicolumn{1}{l}{MoMask w/ UCE} & \cmark & 53.451$^{\pm{1.276}}$ & 14.470$^{\pm{.070}}$ & 15.129$^{\pm{.179}}$ & \underline{5.491}$^{\pm{.000}}$ & \underline{0.148}$^{\pm{.005}}$ & {0.226}$^{\pm{.006}}$ & {0.285}$^{\pm{.008}}$ \\
\multicolumn{1}{l}{MoMask w/ RECE} & \cmark & \underline{13.415}$^{\pm{.439}}$ & \underline{11.205}$^{\pm{.058}}$ & \underline{17.113}$^{\pm{.153}}$ & \textbf{5.138}$^{\pm{.000}}$ & {0.221}$^{\pm{.008}}$ & \underline{0.338}$^{\pm{.007}}$ & \underline{0.424}$^{\pm{.010}}$ \\
\multicolumn{1}{l}{MoMask w/ LCR} & \cmark & \textbf{7.078}$^{\pm.307}$ & \textbf{9.364}$^{\pm.066}$ & 
\textbf{17.167}$^{\pm.203}$ &
3.865$^{\pm.001}$ &
0.317$^{\pm.009}$ & 0.479$^{\pm.009}$ & 0.576$^{\pm.009}$
\\

\midrule \midrule
\multicolumn{1}{l}{BAMM $\mathcal{D}_r$} & \xmark & 59.585$^{\pm{1.552}}$ & 15.913$^{\pm{.087}}$ & 11.352$^{\pm{.228}}$ & 7.985$^{\pm{.165}}$ & 0.084$^{\pm{.004}}$ & 0.148$^{\pm{.006}}$ & 0.205$^{\pm{.007}}$ \\  
\hline 
\multicolumn{1}{l}{{\color[HTML]{\ColorTable} BAMM}} & {\color[HTML]{\ColorTable} --} &
{\color[HTML]{\ColorTable} 56.215$^{\pm{1.799}}$} &
{\color[HTML]{\ColorTable} 14.982$^{\pm{.131}}$} &
{\color[HTML]{\ColorTable} 11.441$^{\pm{.218}}$} &
\color[HTML]{\ColorTable} 7.877$^{\pm{.165}}$ &
{\color[HTML]{\ColorTable} 0.119$^{\pm{.005}}$} &
\color[HTML]{\ColorTable} 0.198$^{\pm{.009}}$ &
\color[HTML]{\ColorTable} 0.261$^{\pm{.009}}$ \\
\multicolumn{1}{l}{ BAMM \textit{FT}}  & \xmark & 
\textbf{54.723}$^{\pm{2.140}}$ &
\textbf{14.798}$^{\pm{.114}}$ &
11.497$^{\pm{.262}}$  &
\textbf{7.968}$^{\pm{.151}}$ &
0.122$^{\pm{.006}}$  &
0.197$^{\pm{.007}}$ &
0.261$^{\pm{.008}}$ \\
\multicolumn{1}{l}{\color[HTML]{\ColorTable} BAMM w/ ESD }  & \xmark & 
50.053$^{\pm{1.824}}$ &
\underline{15.018}$^{\pm{.128}}$ &
\underline{11.424}$^{\pm{.211}}$ &
8.172$^{\pm{.183}}$ &
\textbf{0.105}$^{\pm{.007}}$ &
\textbf{0.179}$^{\pm{.009}}$ &
\textbf{0.248}$^{\pm{.010}}$ \\ 
\multicolumn{1}{l}{BAMM w/ UCE} & \cmark &  67.165$^{\pm{1.614}}$ &  17.469$^{\pm{.099}}$ & 11.801$^{\pm{.307}}$ & 8.397$^{\pm{.170}}$ & 0.051$^{\pm{.004}}$ & 0.095$^{\pm{.005}}$ & 0.138$^{\pm{.005}}$ \\
\multicolumn{1}{l}{BAMM w/ RECE} & \cmark & 67.468$^{\pm{1.383}}$ &  17.800$^{\pm{.077}}$ & 8.967$^{\pm{.340}}$ & 6.759$^{\pm{.146}}$ & 0.019$^{\pm{.002}}$ & 0.038$^{\pm{.004}}$ & 0.055$^{\pm{.006}}$ \\
\multicolumn{1}{l}{BAMM w/ LCR} & \cmark &  \underline{53.639}$^{\pm{1.933}}$ & 15.132$^{\pm{.122}}$ & \textbf{11.402}$^{\pm{.257}}$ & \underline{7.927}$^{\pm{.170}}$ & \underline{0.112}$^{\pm{.005}}$ & \underline{0.192}$^{\pm{.007}}$ & \underline{0.252}$^{\pm{.007}}$ \\ 
\bottomrule
\end{tabular}}
\caption{Forget set full comparison on Motion-X.}
\label{tab:suppmat:mx_forget}

\end{table*}

\begin{table*}
\resizebox{1.\textwidth}{!}{
\begin{tabular}{lc ccccccc}
\toprule
\textit{} &
  \multicolumn{7}{c}{Retain Set} \\ 
  \cmidrule(l{0.4cm}r{0.4cm}){3-9}
\textit{} & Training-free &
  FID $\downarrow$ &
  MM-Dist $\downarrow$ &
  Diversity $\rightarrow$ &
  MultiModality $\uparrow$ &
  \multicolumn{1}{c}{R@1 $\uparrow$} &
  R@2 $\uparrow$ &
  R@3 $\uparrow$ \\ \midrule
\multicolumn{1}{l}{MoMask $\mathcal{D}_r$} & \xmark & 
  11.658$^{\pm.124}$ & 
  9.025$^{\pm.020}$ & 
  19.869$^{\pm.223}$ & 
  4.017$^{\pm.112}$ & 
  0.321$^{\pm.002}$ & 
  0.487$^{\pm.003}$ & 
  0.593$^{\pm.003}$ \\ \hline
\multicolumn{1}{l}{{\color[HTML]{\ColorTable} MoMask}} & {\color[HTML]{\ColorTable} --} & 
  {\color[HTML]{\ColorTable} 3.697$^{\pm.062}$ }& 
  {\color[HTML]{\ColorTable} 8.267$^{\pm.021}$} & 
  {\color[HTML]{\ColorTable} 19.343$^{\pm.177}$} & 
  {\color[HTML]{\ColorTable} 3.757$^{\pm.105}$ }& 
 { \color[HTML]{\ColorTable} 0.384$^{\pm.003}$} & 
  {\color[HTML]{\ColorTable} 0.549$^{\pm.002}$} & 
  {\color[HTML]{\ColorTable} 0.645$^{\pm.002}$}\\
\multicolumn{1}{l}{MoMask \textit{FT}} & \xmark & 
  4.470$^{\pm.046}$ & 
  \underline{8.992}$^{\pm.016}$ & 
  {18.485}$^{\pm.143}$ & 
  3.731$^{\pm.060}$ & 
  \underline{0.337}$^{\pm.002}$ & 
  \underline{0.501}$^{\pm.002}$ & 
  \underline{0.613}$^{\pm.002}$ \\
\multicolumn{1}{l}{MoMask w/ ESD} & \xmark & 
  172.559$^{\pm.535}$ & 
  19.001$^{\pm.014}$ & 
  6.512$^{\pm.076}$ & 
  2.671$^{\pm.076}$ & 
  0.032$^{\pm.001}$ & 
  0.060$^{\pm.001}$ & 
  0.089$^{\pm.002}$ \\
\multicolumn{1}{l}{MoMask w/ UCE} & \cmark & 
  7.252$^{\pm.073}$ & 
  10.843$^{\pm.026}$ & 
  17.950$^{\pm.220}$ & 
  \textbf{4.543}$^{\pm.126}$ & 
  0.275$^{\pm.003}$ & 
  0.412$^{\pm.003}$ & 
  0.498$^{\pm.002}$ \\
\multicolumn{1}{l}{MoMask w/ RECE} & \cmark & 
  \underline{3.689}$^{\pm.056}$ & 
  9.142$^{\pm.019}$ & 
  \underline{19.020}$^{\pm.182}$ & 
  \underline{4.150}$^{\pm.189}$ & 
  0.332$^{\pm.002}$ & 
  0.487$^{\pm.002}$ & 
  0.583$^{\pm.003}$ \\
\multicolumn{1}{l}{MoMask w/ LCR} & \cmark & 
  \textbf{3.658}$^{\pm.060}$ & 
  \textbf{8.329}$^{\pm.017}$ & 
  \textbf{19.344}$^{\pm.184}$ & 
  3.698$^{\pm.136}$ & 
  \textbf{0.381}$^{\pm.003}$ & 
  \textbf{0.541}$^{\pm.003}$ & 
  \textbf{0.639}$^{\pm.001}$ \\ \midrule \midrule
  
\multicolumn{1}{l}{BAMM $\mathcal{D}_r$} & \xmark &
41.739$^{\pm{.457}}$ &
13.918$^{\pm{.030}}$ &
13.689$^{\pm{.163}}$ & 
7.875$^{\pm{.161}}$ &
0.144$^{\pm{.002}}$ &
0.238$^{\pm{.003}}$ &
0.311$^{\pm{.003}}$\\  
\hline
\multicolumn{1}{l}{{\color[HTML]{\ColorTable} BAMM }} & \color[HTML]{\ColorTable} -- &
\color[HTML]{\ColorTable} 54.451$^{\pm{.588}}$ &
\color[HTML]{\ColorTable} 14.526$^{\pm{.033}}$ &
\color[HTML]{\ColorTable} 12.944$^{\pm{.166}}$  &
\color[HTML]{\ColorTable} 8.010$^{\pm{.172}}$ &
\color[HTML]{\ColorTable} 0.135$^{\pm{.002}}$  &
\color[HTML]{\ColorTable} 0.224$^{\pm{.002}}$ &
\color[HTML]{\ColorTable} 0.293$^{\pm{.002}}$ \\
\multicolumn{1}{l}{ BAMM \textit{FT}}  & \xmark &
54.746$^{\pm{.760}}$ &
14.545$^{\pm{.045}}$ &
\textbf{12.900}$^{\pm{.186}}$ &
7.905$^{\pm{.179}}$ &
\textbf{0.136}$^{\pm{.002}}$ &
\textbf{0.225}$^{\pm{.002}}$ &
\textbf{0.294}$^{\pm{.002}}$ \\
\multicolumn{1}{l}{ BAMM w/ ESD }  & \xmark &
\textbf{48.742}$^{\pm{.803}}$ &
\underline{14.377}$^{\pm{.037}}$ &
12.717$^{\pm{.145}}$ &
\textbf{8.447}$^{\pm{.197}}$ &
0.128$^{\pm{.003}}$ &
0.214$^{\pm{.003}}$ &
0.279$^{\pm{.004}}$ \\ 
\multicolumn{1}{l}{BAMM w/ UCE} & \cmark & 88.951$^{\pm{.753}}$ & \textbf{11.314}$^{\pm{.160}}$ & 16.646$^{\pm{.036}}$ & \underline{8.297}$^{\pm{.155}}$ & 0.087$^{\pm{.001}}$ & 0.154$^{\pm{.002}}$ & 0.210$^{\pm{.002}}$ \\
\multicolumn{1}{l}{BAMM w/ RECE} & \cmark & 137.043$^{\pm{.311}}$ & 18.293$^{\pm{.021}}$ & 7.966$^{\pm{.132}}$ & 6.770$^{\pm{.131}}$ & 0.026$^{\pm{.001}}$ & 0.053$^{\pm{.001}}$ & 0.083$^{\pm{.002}}$ \\
\multicolumn{1}{l}{BAMM w/ LCR} & \cmark & \underline{54.481}$^{\pm{.523}}$ & 14.631$^{\pm{.032}}$ & \underline{12.839}$^{\pm{.186}}$ & 8.057$^{\pm{.170}}$ & \underline{0.132}$^{\pm{.001}}$ & \underline{0.217}$^{\pm{.002}}$ & \underline{0.285}$^{\pm{.002}}$ \\ 
\bottomrule
\end{tabular}}
\caption{Retain set full comparison on Motion-X.}
\label{tab:suppmat:mx_retain}
\end{table*}

We present the complete results for the forget and retain sets on the HumanML3D~\cite{humanml3d} dataset, evaluating multimodality and R-precision at top-2 and top-3 for both MoMask~\cite{guo2024momask} and BAMM~\cite{Pinyoanuntapong2024BAMMBA}. Then, we report the corresponding results for the Motion-X~\cite{motionx} dataset.
\paragraph{HumanML3D.}
Table~\ref{tab:suppmat:t2m_forget} presents the expanded results on the forget set of the HumanML3D dataset. LCR consistently outperforms UCE and RECE across all reported metrics for MoMask. While BAMM w/ RECE shows slightly better Diversity and MultiModality scores, LCR achieves a 32.6\% improvement on \MMnotox{} and a 15.8\% increase on R@3.\\
On the retain set, RECE generally achieves better performance on MultiModality, while LCR improves all other metrics across the board.

\paragraph{Motion-X.} Motion-X appears particularly challenging for BAMM~\cite{Pinyoanuntapong2024BAMMBA}.
The FID of our re-trained BAMM is in line with the values reported by their original paper for the HumanML3D dataset. Also, the FID of our retrained MoMask on Motion-X is in line with the values reported by their original paper, just expectedly slightly larger due to the more challenging and noise nature of Motion-X.
The performance of the re-trained BAMM on Motion-X yields a very large FID of 54.451. Since our other re-trainings appear valid, we feel confident of the reported FID value. We believe that, beyond re-training, Motion-X appears to require specific setups which we cannot find documented in the open-source BAMM code. We leave these numbers out of the main paper, as courtesy to authors, and because we believe that better due diligence may be attempted in collaboration with authors, at the earliest authors' availability.
Although notably larger, the figures for LCR applied to BAMM on Motion-X are in line with all previous results and demonstrate that LCR consistently outperforms the other selected baselines. In the forget set (Table~\ref{tab:suppmat:mx_forget}), our model achieves the best \MMnotox{} score, outperforming the second-best technique by 9.1\% with MoMask and 13.38\% with BAMM. Additionally, BAMM w/ LCR achieves the best performance on retrieval metrics.
In the retain set (Table~\ref{tab:suppmat:mx_retain}), LCR improves FID by 28.9\% and 38.7\% when applied to MoMask and BAMM, respectively. UCE achieves the best \MMnotox{} score, while LCR performs best on Diversity and R@3, with a 7.5\% improvement.

\section{Implementation Details}

\paragraph{ESD.} 
We adapt the ESD loss~\cite{Gandikota2023ErasingCF} to MoMask and BAMM.
While ESD is originally designed for diffusion models and operates by optimizing the predicted noise, MoMask and BAMM are not diffusion-based.
Instead, we apply the loss directly to the latent representation produced by the masked transformer, which is shared by both architectures.
We use the default guidance scale of $\eta = 3$, as in the original ESD setup.
In their setting, ESD optimizes a single concept over 200 iterations. To ensure a fair comparison, we evaluate using the same evaluation set adopted in our benchmark.
However, since this test set consists of approximately $1000$ samples representing the same concepts through diverse prompts, we opt not to repeat optimization over the same prompt multiple times.

\paragraph{UCE and RECE.}
The algorithms from UCE~\cite{gandikota2024uce} and RECE~\cite{gong2024rece} have been readapted for human motion unlearning. Originally, these methods operated on the cross-attention mechanism between text and motion embeddings. However, since the T2M models MoMask~\cite{guo2024momask} and BAMM~\cite{Pinyoanuntapong2024BAMMBA} do not incorporate cross-attention, we instead apply the method to the projection embedding that maps the CLIP embedding to the motion embedding (\texttt{.cond\_emb}), following the approach that the authors of UCE use to adapt it to the FLUX architecture in \citet{gandikota2024uce_flux}.

\paragraph{Motion-X.}
Motion-X~\cite{motionx} provides SMPL-X~\cite{SMPL-X:2019} motion data, which includes detailed hand and facial features. To align this dataset with our setup, we first process it using the official code provided by its authors
to convert SMPL-X motion features into SMPL~\cite{SMPL:2015} representations.

Next, to ensure consistency with HumanML3D~\cite{humanml3d}, we preprocess the text prompts using the Semantic Role Labeling (SRL) tool provided by the HumanML3D authors.
Some manual refinements are applied due to variations in the formatting of prompts within the Motion-X dataset. 

Additionally, since Motion-X is organized as multiple datasets, we restructure it into a format that closely resembles HumanML3D by flattening its structure. 
We will release this newly processed dataset to the community, appropriately crediting the original authors.\\
For evaluation purposes, the feature extration has been trained following \cite{humanml3d}'s implementation
, for 300 epochs.

\paragraph{Training.}
We train MoMask and BAMM on Motion-X and on the clean $\mathcal{D}_r$ splits of HumanML3D and Motion-X. We run up to 50 epochs for the VQ-VAE and up to 500 epochs for the Masked and Residual Transformers, stopping training once the loss plateaus.

For MoMask, we train the entire pipeline, composed by the VQ-VAE, the masked transformer, and residual transformer. For BAMM, which builds upon MoMask by modifying only the masked transformer, we do not retrain the residual transformer, as this is also restricted by the original code implementation.
Our training is conducted on a single NVIDIA A100 GPU. On Motion-X, training all stages of MoMask takes approximately 34 hours. Training from scratch on the cleaned versions of HumanML3D and Motion-X takes approximately one day for each dataset.

When finetuning our models, we do it for five additional epochs, requiring about 10 minutes in total for both the masked and residual transformers. Our LCR method takes approximately 15 seconds.

\paragraph{Fine Tuning Epochs} Table \ref{tab:reb:ft} presents an ablation on the number of fine-tuning epochs for MoMask.
We observe that longer fine-tuning shifts the model’s behavior from MoMask toward MoMask $\mathcal{D}_r$, effectively reducing violent motions.
However, a full training requires 500 epochs and takes approximately 24 hours. This level of tuning only slightly outperforms LCR, which, by contrast, completes in just 15 seconds.
Therefore, although extended fine-tuning is effective in suppressing violent content, it is not a practical solution.

\begin{table}[!ht]
\resizebox{\linewidth}{!}{
\begin{tabular}{lccccc}
\toprule
& \multicolumn{2}{c}{Forget Set} & \multicolumn{2}{c}{Retain Set} &  \\ 
\cmidrule(lr){2-3} \cmidrule(lr){4-5}
\textit{} & FID $\rightarrow$ & \MMnotox $\downarrow$ & FID $\downarrow$ & MM-Dist$\downarrow$ & Time \\ 
\midrule
\multicolumn{1}{l}{MoMask $\mathcal{D}_r$} &
    13.644$^{\pm{.365}}$ &
    4.392$^{\pm{.041}}$ &
    0.093$^{\pm{.003}}$ &
    2.700$^{\pm{.007}}$ & 
    35h\\
\midrule
\multicolumn{1}{l}{{\color[HTML]{828282} MoMask}} &
  {\color[HTML]{828282} 2.028$^{\pm{.127}}$} &
  {\color[HTML]{828282} 9.001$^{\pm{.071}}$} &
  {\color[HTML]{828282} 2.686$^{\pm{.045}}$} &
  {\color[HTML]{828282} 8.076$^{\pm{.017}}$} \\ 
  
\multicolumn{1}{l}{MoMask FT +5} &
    2.295$^{\pm{.065}}$ &
    4.497$^{\pm{.018}}$ &
    0.070$^{\pm{.001}}$ &
    3.034$^{\pm{.003}}$ &
    10m \\
 \multicolumn{1}{l}{MoMask FT +500} &
     5.920$^{\pm{.168}}$ &
     4.662$^{\pm{.035}}$ &
     0.055$^{\pm{.002}}$ &
     2.714$^{\pm{.008}}$ &
     24h \\
\bottomrule

\end{tabular}}
\caption{Ablation on HumanML3D of fine-tuning epochs.}
\label{tab:reb:ft}
\end{table}

\section{Ablations on UCE/RECE}\label{sec:uce_ablation}

\paragraph{Preserve Scale.}
We ablate the $\lambda$ hyperparameter in UCE~\cite{gandikota2024uce} and RECE, which controls the trade-off between unlearning efficacy and knowledge preservation. A high $\lambda$ enforces strict retention of original capabilities, while low values permit greater deviation to achieve unlearning.
Tables~\ref{tab:uce_ablation} and~\ref{tab:rece_ablation} reveal a clear dichotomy. At high $\lambda$, models preserve excellent retain performance (FID and R@1 nearly unchanged from the original model), but fail to unlearn: forget set FID remains excessively low, meaning the model still generates realistic toxic motions similar to the training data.
At low $\lambda$, unlearning succeeds: forget set FID increases substantially (higher FID indicating the model no longer reproduces toxic motions), approaching the retrained baseline. However, this causes catastrophic forgetting: retain FID rises (worse quality) and R@1 falls (worse text-motion alignment), indicating degraded performance on safe motions.
These results confirm that $\lambda$ requires careful tuning to balance forgetting toxic concepts against preserving model utility on safe behaviors. We set $\lambda=0.5$ to optimize this trade-off.

\begin{table}[ht]
\resizebox{\linewidth}{!}{
\begin{tabular}{lccccc}
\toprule
& & \multicolumn{4}{c}{Forget Set}             \\ \cmidrule{3-6} 
& \textit{$\lambda$} & FID $\rightarrow$   & MM-Safe $\downarrow$ & Diversity $\rightarrow$ & R@1 $\rightarrow$    \\ \midrule
MoMask $\mathcal{D}_r$ & - & 16.358 & 4.497       & 6.955       & 0.118 \\ \hline
\multirow{8}{*}{UCE} & 2 & 1.157 & 5.109       & 5.632       & 0.180 \\
 & 1.5 & 11.937       & 4.642 & \textbf{7.079}    & 0.137         \\
 & 1 & 1.199 & 5.096       & 5.553       & 0.186 \\
 & 0.75 & 12.008 & 4.638       & 7.114       & 0.135 \\ 
 & 0.5 & 11.860 & \textbf{4.626}       & 7.144       & 0.135 \\ 
 & 0.1 & 12.198 & 4.639       & 7.143       & 0.134 \\ 
 & 0.05 & 12.431 & 4.638       & 7.116       & 0.134 \\ 
 & 0.01 & \textbf{14.556} & 4.678       & 7.323       & \textbf{0.127} \\ \midrule \toprule
& & \multicolumn{4}{c}{Retain Set}             \\ \cmidrule{3-6}
 & \textit{$\lambda$} & FID $\downarrow$   & MM-Dist $\downarrow$ & Diversity $\rightarrow$ & R@1 $\uparrow$    \\ \midrule
MoMask $\mathcal{D}_r$ & - & 0.075   & 2.959       & 9.545       & 0.512 \\ 
\hline
\multirow{8}{*}{UCE} & 2 & \textbf{0.041}   & \textbf{2.929}       & 9.658      & 0.519 \\
 & 1.5 & 0.089 & 3.099 & 9.703       & 0.497 \\
 & 1 & \textbf{0.041}   & 2.930       & \textbf{9.640}       & \textbf{0.520} \\
 & 0.75 & 0.090   & 3.103       & 9.716    & 0.496 \\
 & 0.5 & 0.090         & 3.100          & 9.733 & 0.497         \\ 
 & 0.1 & 0.097         & 3.105          & 9.753 & 0.496         \\
 & 0.05 & 0.102         & 3.104          & 9.690 & 0.495         \\
 & 0.01 & 0.125         & 3.121          & 9.680 & 0.492         \\\bottomrule
\end{tabular}}
\caption{Ablation on the preserve scale $\lambda$ on UCE.}
\label{tab:uce_ablation}
\end{table}

\begin{table}[ht]
\resizebox{\linewidth}{!}{
\begin{tabular}{llccccc}
\toprule
& & & \multicolumn{4}{c}{Forget Set}             \\ \cmidrule{4-7} 
& \textit{Epochs} & \textit{$\lambda$} & FID $\rightarrow$   & MM-Safe $\downarrow$ & Diversity $\rightarrow$ & R@1 $\rightarrow$    \\ \midrule
 $\mathcal{D}_r$ & - & - & 16.358 & 4.497       & 6.955       & 0.118 \\ \hline
\multirow{10}{*}{RECE} & 1 & 2 & 1.176 & 5.101       & 5.625       & 0.181 \\
 & 1 & 1.5 & 6.977       & 4.871 & 6.587    & 0.156         \\
 & 1 & 1 & 1.167 & 5.087       & 5.579       & 0.186 \\
 & 1 & 0.75 & 7.058 & \textbf{4.868}       & 6.571       & 0.150 \\ 
 & 1 & 0.5 & 6.952 & 4.899       & 6.548       & 0.148 \\ 
 & 1 & 0.1 & 6.931 & 4.901       & 6.535       & 0.153 \\ 
 & 1 & 0.05 & 7.056 & 4.883       & 6.473       & 0.152 \\ 
 & 1 & 0.01 & 9.467 & 4.890       & \textbf{6.865}       & 0.148 \\ 
 & 3 & 0.05 & \textbf{20.596} & 5.381       & 8.318       & 0.133 \\ 
 &3 & 0.01 & 20.292 & 5.309       & 8.267       & \textbf{0.128} \\ \midrule \toprule
& & & \multicolumn{4}{c}{Retain Set}             \\ \cmidrule{4-7}
 & \textit{Epochs} & \textit{$\lambda$} & FID $\downarrow$   & MM-Dist $\downarrow$ & Diversity $\rightarrow$ & R@1 $\uparrow$    \\ \midrule
 $\mathcal{D}_r$ & - & - & 0.075   & 2.959       & 9.545       & 0.512 \\ 
\hline
\multirow{10}{*}{RECE} & 1 & 2 & 0.041   & 2.932       & 9.664      & 0.518 \\
 &1 & 1.5 & 0.140 & 3.120 & 9.785       & 0.495 \\
 &1 & 1 & \textbf{0.041}   & \textbf{2.931}       & \textbf{9.611}       & \textbf{0.521} \\
 &1 & 0.75 & 0.144   & 3.124       & 9.739    & 0.492 \\
 &1 & 0.5 & 0.144         & 3.124          & 9.814 & 0.493         \\ 
 &1 & 0.1 & 0.146         & 3.126          & 9.860 & 0.493         \\
 & 1& 0.05 & 0.153         & 3.131          & 9.796 & 0.493         \\
 & 1& 0.01 & 0.196         & 3.151          & 9.713 & 0.491         \\
 &3 & 0.05 & 0.514         & 3.566          & 10.047 & 0.434         \\
 &3 & 0.01 & 0.484         & 3.508          & 10.036 & 0.443         \\\bottomrule
\end{tabular}}
\caption{Ablation on strength of preserve scale $\lambda$ and number of epochs in RECE.}
\label{tab:rece_ablation}
\end{table}

\paragraph{Target Keyword.} When adapting UCE~\cite{gandikota2024uce} and RECE~\cite{gong2024rece} for human motion unlearning, it is necessary to predefine the target keyword the toxic motions will be mapped to. We experiment with three different keywords: ``\textit{walk}", ``\textit{stand}", and an empty string.
Walking and standing are among the most frequent actions in the dataset and often serve foundations for more complex motions.
The keyword ``\textit{stand}" appears in various prompts, referring to different nuances of the action, such as ``\textit{standing still}" or ``\textit{standing up}". Similarly, the keyword ``\textit{walk}" may correspond to different directional variations, like ``\textit{walking left}" or ``\textit{walking right}". 
The empty string, on the other hand, provides greater flexibility by allowing the model to remain less constrained by a predefined target, and aligns with the target chosen in \cite{gandikota2024uce, gong2024rece}.
Our results in Table \ref{tab:uce_rece_comparison} indicate that using the empty string generally leads to better performance. Therefore, we adopt it as the primary choice in the main.

Our method, LCR, operates on the discrete latent space without specifying any a priori target word.

\begin{table*}[!b]
\centering
\resizebox{1.\textwidth}{!}{
\begin{tabular}{ll cccc cccc}
\toprule
 &
   &
  \multicolumn{4}{c }{Forget Set} &
  \multicolumn{4}{c}{Retain Set} \\ \cmidrule(l{0.4cm}r{0.4cm}){3-6} \cmidrule(l{0.4cm}r{0.4cm}){7-10} 
 &
   &
  FID$\rightarrow$ &
  \MMnotox{} $\downarrow$ &
  Diversity$\rightarrow$ &
  R@1$\rightarrow$ &
  FID$\downarrow$ &
  MM-Dist $\downarrow$ &
  Diversity$\rightarrow$ &
  R@1$\uparrow$ \\ 
  \midrule
MoMask $\mathcal{D}_r$ &  & 16.358$^{\pm.150}$ & 4.497$^{\pm.018}$ & 6.955$^{\pm.054}$ & 0.118$^{\pm.005}$ & 0.075$^{\pm.001}$ & 2.959$^{\pm.002}$ & 9.545$^{\pm.086}$ & 0.512$^{\pm.001}$ \\ 
\midrule
\multicolumn{1}{l}{\multirow{3}{*}{MoMask w/ UCE}} &
  ``\textit{Stand}" &
  43.660$^{\pm{.516}}$ &
  6.350$^{\pm{.034}}$ &
  {9.173}$^{\pm{.083}}$ &
  \textbf{0.110}$^{\pm{.004}}$ &
  1.379$^{\pm{.027}}$ &
  3.653$^{\pm{.008}}$ &
  10.178$^{\pm{.113}}$ &
  0.233$^{\pm{.001}}$ \\
\multicolumn{1}{l}{} &
  ``\textit{Walk}" &
  75.764$^{\pm{.823}}$ &
  7.930$^{\pm{.047}}$ &
  \textbf{6.980}$^{\pm{.070}}$ &
  0.090$^{\pm{.004}}$ &
  10.851$^{\pm{.079}}$ &
  4.910$^{\pm{.010}}$ &
  8.698$^{\pm{.108}}$ &
  0.193$^{\pm{.001}}$ \\
 &
  `` " &
  \textbf{11.860}$^{\pm.154}$ & \textbf{4.626}$^{\pm.013}$ & 7.144$^{\pm.080}$ & 0.135$^{\pm.008}$ & \textbf{0.090}$^{\pm.001}$ & \textbf{3.100}$^{\pm.003}$ & \textbf{9.733}$^{\pm.089}$ & \textbf{0.497}$^{\pm.001}$ \\ \midrule
\multicolumn{1}{l}{\multirow{3}{*}{MoMask w/ RECE}} &
  ``\textit{Stand}" &
  75.091$^{\pm{.957}}$ &
  9.037$^{\pm{.063}}$ &
  \textbf{8.765}$^{\pm{.107}}$ &
  0.058$^{\pm{.003}}$ &
  {5.780}$^{\pm{.029}}$ &
  {4.482}$^{\pm{.007}}$ &
  \textbf{9.518}$^{\pm{.137}}$ &
  0.195$^{\pm{.001}}$ \\
\multicolumn{1}{l}{} &
  ``\textit{Walk}" &
  120.591$^{\pm{.1.092}}$ &
  10.599$^{\pm{.038}}$ &
  4.926$^{\pm{.102}}$ &
  0.039$^{\pm{.002}}$ &
  49.341$^{\pm{.226}}$ &
  8.953$^{\pm{.013}}$ &
  6.339$^{\pm{.117}}$ &
  0.066$^{\pm{.0007}}$ \\
& `` " &
 \textbf{6.952}$^{\pm.110}$ & \textbf{4.899}$^{\pm.016}$ & \textbf{6.548}$^{\pm.048}$ & \textbf{0.148}$^{\pm.006}$ & \textbf{0.144}$^{\pm.002}$ & \textbf{3.124}$^{\pm.004}$ & 9.814$^{\pm.099}$ & \textbf{0.493}$^{\pm.001}$ \\ \bottomrule

\end{tabular}}
\caption{Comparison of UCE and RECE across different target concepts, evaluated on HumanML3D.
}
\label{tab:uce_rece_comparison}
\end{table*}

\section{Metrics}
\label{sec:loss_metr}

We provide the mathematical formulations for evaluating the generated results based on established literature \cite{humanml3d, chen2023mld, zhang2023t2mgpt, tevet2023mdm, jiang2023motiongpt, Pinyoanuntapong2024BAMMBA, guo2024momask, sampieri24ladiff}.

\paragraph{R-Precision and \MMnotox{}.}
In text-to-motion tasks, \cite{humanml3d} introduces motion and text feature extractors that produce geometrically aligned representations for paired text-motion samples. 
Within this feature space, we compute R-Precision by embedding generated motions alongside mismatched samples and measuring the top-1/2/3 retrieval accuracy.

To assess semantic alignment between generated motions and their text prompts, we compute the Multi-modal Distance (MM-Dist), which captures how well a motion matches its conditioning description.
For the forget set $\mathcal{D}_f$, we use a variant called Multi-modal Safe Distance (\MMnotox{}), which modifies MM-Dist by censoring violent keywords in the text prior to computing the distance.
This ensures that both violent and nonsensical motions result in low alignment with the blanked-out prompt, allowing us to better distinguish between models that degrade motion quality and those that selectively unlearn while preserving coherence in non-violent behavior.
This distinction is particularly important in motion generation, where outputs are composed of multiple poses and sub-motions, and degradation can be subtle and localized.

R-Precision and Multi-modal distances are computed using the $L_2$ distance.

\paragraph{FID.} Motion quality is evaluated using the Fréchet Inception Distance (FID), which compares the statistical similarity between the distributions of generated and real motions. This is achieved by computing the L2-loss between their latent feature representations.

\paragraph{Diversity and MultiModality.} Diversity and MultiModality metrics are employed to measure both the overall variability in generated motions and the diversity of motions corresponding to the same text prompt.

Diversity is computed by randomly splitting the generated motion set into two equal-sized subsets, $\{m_1, \dots, m_{\mathcal{M}_d}\}$ and $\{m'_1, \dots, m'_{\mathcal{M}_d}\}$, each containing motion feature vectors. It is then defined as:

\begin{equation}
    \text{Diversity} = \frac{1}{\mathcal{M}_d} \sum_{i=1}^{\mathcal{M}_d} ||m_i - m'_{i}||.
\end{equation}

MultiModality is assessed by first selecting $T_d$ random text descriptions from the dataset. For each text description, we sample two equal-sized subsets of motion feature vectors, denoted as $\{m_{t,1}, \dots, m_{t,\mathcal{M}_d}\}$ and $\{m'_{t,1}, \dots, m'_{t,\mathcal{M}_d}\}$. MultiModality is then computed as:

\begin{equation}
    \text{MultiModality} = \frac{1}{\mathcal{T}_d \cdot \mathcal{M}_d} \sum_{t=1}^{\mathcal{T}_d} \sum_{i=1}^{\mathcal{M}_d} ||m_{t,i} - m'_{t,i}||.
\end{equation}

\end{document}